\pdfoutput=1

\documentclass[11pt]{article}

\usepackage[]{ACL2023}

\usepackage{times}
\usepackage{latexsym}
\usepackage{booktabs}
\usepackage{tikz}
\usepackage{siunitx}
\usepackage{tabularx}
\usepackage{graphicx}
\usepackage[T1]{fontenc}
\usepackage[utf8]{inputenc}
\usepackage{microtype}
\usepackage{pifont}
\usepackage{inconsolata}
\usepackage{amsmath} 
\usepackage{makecell}
\usepackage{xspace}
\usepackage{xfp}
\usepackage{xcolor}

\newcommand{\ms}[2]{%
  \makecell[c]{#1\\[-0.1ex]\smash{\textcolor{gray}{\scriptsize(#2)}}}%
}

\newcommand{\msb}[2]{%
  \makecell[c]{\textbf{#1}\\[-0.1ex]\smash{\textcolor{gray}{\scriptsize(#2)}}}%
}

\newcommand{\spavgfull}{\textsc{Paragram-SP}\xspace}
\newcommand{\semsim}{\textsc{SemSim}\xspace}
\newcommand{\ninstances}{\textsc{\#instn}\xspace}
\newcommand{\nsources}{\textsc{\#src}\xspace}

\definecolor{barblue}{rgb}{0.70,0.82,0.92}
\definecolor{barorange}{rgb}{0.95,0.75,0.50}
\definecolor{humanblue}{rgb}{0.82,0.90,0.97}
\definecolor{syntheticblue}{rgb}{0.65,0.775,0.89}

\newcommand{\bluebarcell}[2]{%
  \begin{tikzpicture}[baseline=-0.6ex]
    \fill[humanblue] (0,0) rectangle ({\fpeval{1.0*(#1)/(#2)}},0.32);
    \node[anchor=west, font=\scriptsize, text=black] at (0.08,0.16) {\num{#1}};
  \end{tikzpicture}%
}

\newcommand{\orangebarcell}[2]{%
  \begin{tikzpicture}[baseline=-0.6ex]
    \fill[barorange] (0,0) rectangle ({\fpeval{1.0*(#1)/(#2)}},0.32);
    \node[anchor=west, font=\scriptsize, text=black] at (0.08,0.16) {\num{#1}};
  \end{tikzpicture}%
}

\newcommand{\stackedbarcell}[3]{%
  \begin{tikzpicture}[baseline=-0.6ex]

    \fill[humanblue] (0,0)
      rectangle ({\fpeval{1.0*(#1)/(#3)}},0.32);

    \fill[syntheticblue] ({\fpeval{1.0*(#1)/(#3)}},0)
      rectangle ({\fpeval{1.0*(#2)/(#3)}},0.32);

    \node[anchor=west, font=\scriptsize, text=black]
      at (0.08,0.16) {\num{#2}};
  \end{tikzpicture}%
}

\title{Learning from Many Voices: \\ Literary MT Using Multi-Reference Human and Synthetic Data}

\author{Si Wu \\ Northeastern University \\  \texttt{siwu@ccs.neu.edu}
\And
John Wieting \\ Google DeepMind\\
\texttt{jwieting@google.com}
\And
David A. Smith \\ Northeastern University \\ \texttt{dasmith@ccs.neu.edu}}

\begin{document}
\maketitle
\begin{abstract}

Unlike many other texts, literary works are often translated multiple times. We investigate strategies for leveraging these multi-reference datasets to improve literary machine translation. We propose a filtering framework based on semantic similarity to identify source texts whose references display meaningful variation while remaining faithful. We find that fine-tuning with medium to high semantic similarity data substantially outperforms low semantic similarity data. Moreover, using medium and high semantic similarity data achieves comparable or better performance than using the full unfiltered data.

Synthetic translations generated by LLMs are economical and convenient alternatives to human expert translations; however, we find fine-tuning on human expert translations outperforms fine-tuning on synthetically augmented data in automatic metrics and human evaluations, demonstrating the indispensable value of human expert translations for fine-tuning literary machine translation models.

\end{abstract}

\section{Introduction}

\begin{figure}[t]
\centering
\includegraphics[width=0.95\linewidth]{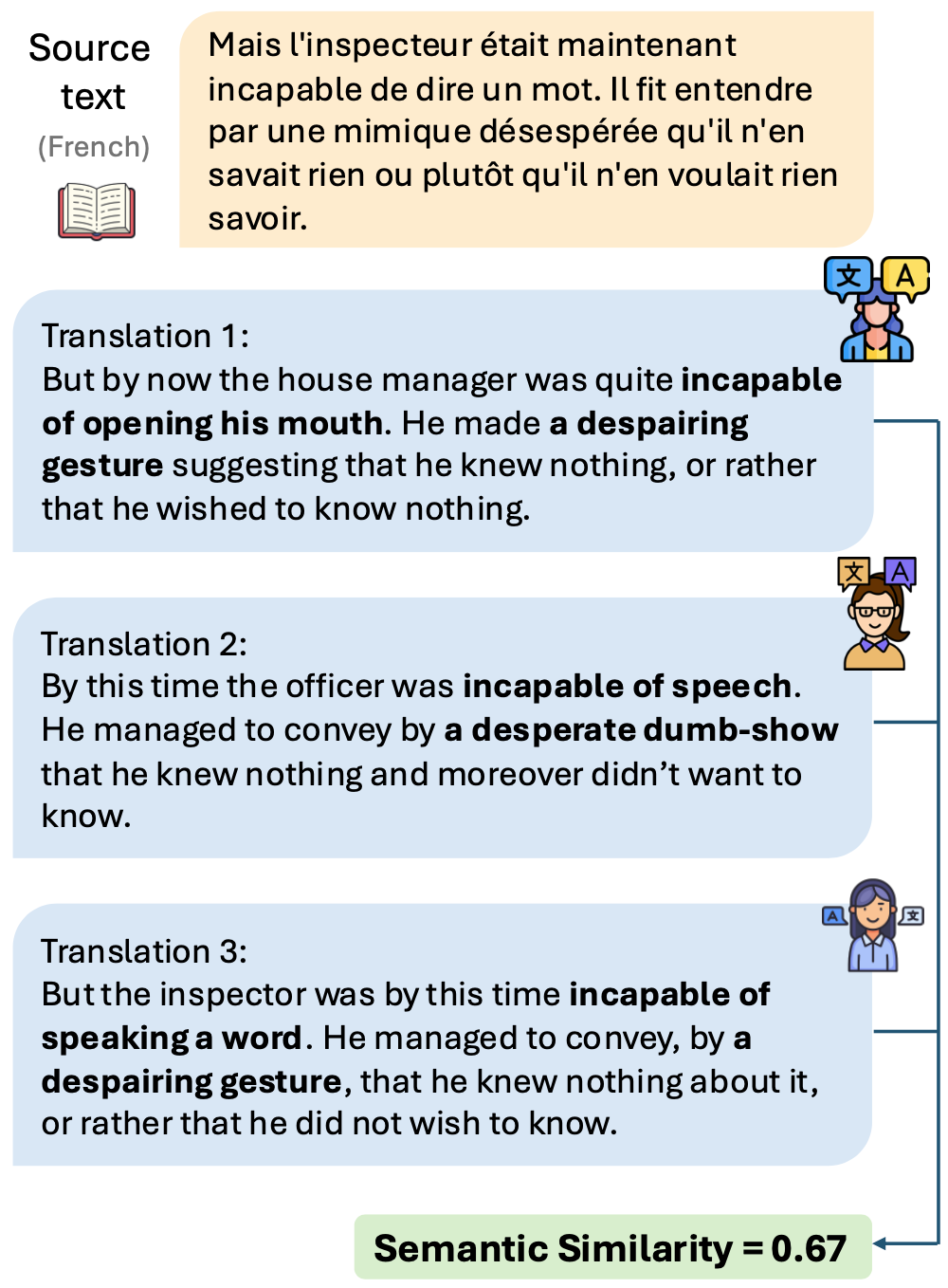}
\caption{An example of semantic variation across human translations of the same source text. We analyze a paragraph from \emph{Le Fantôme de l'Opéra} (\textit{The Phantom of the Opera}) by Gaston Leroux. The original French paragraph was translated into English by three different literary translators. Using our proposed scoring system \semsim, the average pairwise semantic similarity across the English translations is 0.67.}
\label{fig-first-page-example}
\end{figure}

Literary translation presents unique challenges for machine translation (MT). Literary works are deeply embedded in the source language and culture; therefore, it can be difficult to reproduce the same meaning while preserving cultural context and stylistic nuance in another language. Moreover, literary translations are meant to be read. Fully explicating all interpretations would require substantial annotation or commentary (e.g., in footnotes)---what Nabokov memorably called ``translations with copious footnotes reaching up like skyscrapers'' \citep{nabokov1955problems}---and reduce readability.

As such, literary translators must balance multiple aspects of a source text when translating: meaning, form, rhyme, style, and cultural context \citep{jakobson1987linguistics,andre, hofstadter1997music}. Some emphasize lexical and syntactic imitation (formal equivalence), while others sacrifice form for comprehensibility (dynamic/functional equivalence) \citep{nida1964toward}. Read together, different translations of the same source text collectively provide diverse interpretations and a richer understanding of the original work. For example, in Figure 1, the three translations vary in style and word choice, yet together they reveal not only the meaning and literary style of the French text, but also some of its etymological associations: translation 2's ``dumb-show'' draws out the theatrical associations of the French word \textit{mimique}, which is related to mime and gesture-based expression.

Prior work has explored methods to synthetically generate additional references for a source text via back-translation or paraphrasing \citep{madnani-etal-2008-multiple, zheng-etal-2018-multi, khayrallah-etal-2020-simulated}. However, these works are conducted within non-literary domains such as news, and the text is usually at sentence-level. News and literary texts differ fundamentally in purpose and style: news is generally written to be factual, direct, and unambiguous, whereas literary writing is sometimes intentionally open to interpretation, with authors using devices such as metaphor, imagery, symbolism, rhythm, and rhyme. Due to the very nature of literary text, scholars and readers in comparative literature and translation studies often compare multiple translations of a source text to deepen their understanding of the original and of how its meanings are rendered across languages \citep{rose1997translation, stenberg2018multiplicitypoetry, szymanska2026manifesto}.

Moreover, well-known literary works often have multiple translations, particularly in high-resource language pairs. It is reasonable to leverage the wealth of expert-produced translations that are already available. However, the question is how to effectively utilize them when training or fine-tuning an MT system. Additionally, it's worth investigating the performance gap between models trained on expert translations and those trained on synthetic paraphrases. Synthetic alternatives are more economical and may be particularly beneficial to low-resource languages where parallel data are scarce \citep{khayrallah-etal-2020-simulated, de-gibert-etal-2025-scaling}.

In this paper, we investigate how to leverage multiple references for literary translation. Specifically, we propose a filtering approach based on semantic similarity scores: selecting source texts whose reference translations collectively preserve semantic fidelity while exhibiting meaningful lexical and syntactic variation, as measured by pairwise semantic similarity scores.

This paper examines three hypotheses:

    \paragraph{H1. Different ranges of semantic similarity will lead to different performance:} High and medium \semsim lead to substantially better translation than low \semsim. (Section~\ref{Section-finding-optimal})

    \paragraph{H2. A filtered dataset is better for fine-tuning literary MT than an unfiltered dataset:} Combining high and medium \semsim data achieves MT performance comparable or better than unfiltered dataset. (Section~\ref{section-combine-ranges})

    \paragraph{H3. A synthetic-augmented multi-reference dataset will perform comparably to an expert-translated dataset:} Synthetic augmented dataset does not improve MT performance as reliably as human expert-translated multi-reference data and can underperform an unaugmented single-reference dataset, depending on the fine-tuning models used for augmentation. (Section~\ref{Section-synthetic})

For all the hypotheses, we compare performance measured by automatic metrics and evaluations from human translators. We also present zero-shot MT results from commercial and open-weight instruction-tuned models in Appendix~\ref{section-zero-shot}, and provide example outputs in Appendix~\ref{appendix-example-outputs}.

\section{Related work}

Literary machine translation is known to be a challenging task \citep{taivalkoski-shilov-2019-free, brusasco2022pragmatic, human-vs-NMT-paster}. Unlike news and academic writing, literary language often deploys devices such as imagery, symbolism, metaphor, rhythm, and rhyme. Literary text is deeply embedded in its linguistic, cultural and historical contexts. Polysemy is one example: when multiple meanings of a word in the source text are intended to be evoked, but in the target language there is no word with the same one-to-many mapping, there is no direct translation. In this situation, a translator will have to decide what to prioritize in a translation. As such, to deepen their understanding of the linguistic and cultural nuance of the source text, scholars in comparative literature and translation studies often compare multiple translations \citep{rose1997translation, stenberg2018multiplicitypoetry, szymanska2026manifesto}.

In machine translation, the idea of using multiple target translations of a source text to improve understanding of the source-target language pair has long been explored in various ways. Much of this work is through synthetically creating new target texts: from using pivot-based paraphrasers \citep{callison-burch-etal-2006-improved, madnani-etal-2008-multiple}, to back-translation in statistical and neural machine translation \citep{sennrich-etal-2016-improving, poncelas-etal-2018-investigating, edunov-etal-2018-understanding}. More recently, LLMs have been used for synthetic data generation in many different tasks \citep{long-etal-2024-llms}, and LLMs' ability to generate high-quality text is particularly beneficial in low-resource settings \citep{scalvini-etal-2025-rethinking, de-gibert-etal-2025-scaling}. Although LLMs have improved the quality and complexity of paraphrase generation, they still make mistakes, particularly for outputs longer than a single sentence \citep{Jayawardena_2024, Jayawardena_2024_gen, su2022read, horvitz2024paraguide, mishra2024finegrained}. Most prior work focuses on non-literary domains and primarily at sentence-level. 

Many have explored employing post-editing to improve literary MT \citep{besacier-schwartz-2015-automated, thai-etal-2022-exploring, castaldo2025extendingcreamtleveraginglarge}. While some have looked into fine-tuning methods for the literary MT task \citep{li-etal-2024-linchance, mikelenic-etal-2025-fine}, we are unaware of work on how to optimally fine-tune MT model for the literary domain, particularly with multiple translations per source text. Moreover, no prior work has directly compared the effects of fine-tuning on human data vs. synthetic data for literary MT.

Finally, our filtering criterion differs from the general notion of ``noise'' in MT: we filter based on semantic similarity of target texts of the same source text, not the mismatches or semantic divergence between source and target. Nevertheless, prior work on data filtering shows that training data quality matters for MT performance \citep{carpuat-etal-2017-detecting, khayrallah-koehn-2018-impact, briakou-carpuat-2021-beyond, steingrimsson-etal-2023-filtering}.

\section{Experimental setup} \label{sec-exp-setup}
We describe the models we use for MT and automatic metrics we use in Section~\ref{section-models-description}, and the details of pre-processing steps in Appendix~\ref{pre-processing}. We discuss each hypothesis and its results using the multi-reference datasets produced from the pre-processing step in Sections~\ref{Section-finding-optimal} to \ref{Section-synthetic}.

\subsection{Literary Translation Data}
Popular works of world literature are often translated into English by multiple translators, and these translations can differ stylistically while remaining faithful to the original text. The Par3 dataset aligns multiple English translations of each non-English book at the paragraph level: each source paragraph has at least two distinct English translations produced by expert literary translators \citep{thai-etal-2022-exploring}. Translating literature at paragraph-level instead of sentence-level incorporates more discourse features and has been proven to improve literary MT \citep{voigt-jurafsky-2012-towards}. Although Par3 covers 19 source languages, we conduct experiments only on the French-English and Russian-English language pairs, which account for 48\% and 32\% of the total source paragraphs, respectively. Although both language pairs exhibit similar trends in automatic metrics, our discussion in Sections~\ref{Section-finding-optimal} to \ref{Section-synthetic} will mainly focus on French-English, since we only conducted human evaluations on this language pair.

\subsection{Models and Automatic Metrics} \label{section-models-description}
We fine-tune three models for the literary machine translation task: mT5-Large \citep{xue-etal-2021-mt5}, Gemma3-12B-PT \citep{gemmateam2025gemma3technicalreport}, and OpusMT \citep{tiedemann2023democratizing, TiedemannThottingal:EAMT2020}, and we use BLEU as the early-stopping criterion for all models. Detailed model descriptions, training details, and the descriptions of automatic metrics are in Appendix~\ref{appendix-model-metrics}.

\section{Measuring Semantic Similarity Among References of the Same Source Text} \label{subsec-semsim}

Intuitively, reference translations of the same source text should be semantically similar but not identical: if two references' semantic similarity is too high, the meaning of the two would be nearly identical, and the additional reference would provide little new learning signal; conversely, if the semantic similarity is too low, the discrepancy may indicate noise, such as source--target misalignment or overly divergent translator interpretations.

We measure semantic similarity between reference translations of the same source text using \semsim. 
This similarity score allows us to quantify the variations among different references in lexical choice or syntactic structure while expressing the same underlying meaning.

\semsim utilizes a semantic similarity scoring model \spavgfull \citep{wieting-etal-2022-paraphrastic}. \spavgfull score ranges from -1 to 1, where a lower score indicates low semantic similarity, and a higher score indicates high semantic similarity. A model similar to this one from~\citet{wieting-etal-2019-simple} was successfully used as a semantic similarity reward in minimum risk training for neural machine translation~\citep{wieting-etal-2019-beyond}, building on prior minimum risk training methods for MT 
\citep{smith06:minrisk,shen-etal-2016-minimum}. Since \spavgfull can only score the semantic similarity for two texts, with more than two translations, we take the average pairwise \spavgfull score as the source text's \semsim. Mathematically:

$$\textit{\semsim (s)} = \frac{1}{|C|} \sum_{(a,b)\in C} \text{\spavgfull}(a,b) $$ 
where $M$ is the set of human translations given a source text $s$, and $C$ is all possible combinations of a set of pairs in $M$. \spavgfull$(a,b)$ is the \spavgfull score given a pair of English translations $a$ and $b$.

Note that \spavgfull computes similarity over embeddings and does not explicitly penalize length. There is an implicit effect only insofar as a reference that omits or heavily alters content will differ in embedding space and score lower, but length alone is not the driver.

We compute \semsim for each source text and use these scores to filter and create new datasets. Notice that \semsim is computed only with a source text's English translations without relying on the source text itself. 

We assume source-target alignment is verified during preprocessing, e.g., Par3 verified source-target alignment. The alignment method used in preprocessing is itself an active research area, but we believe the proposed \semsim score helps further filter out misaligned examples regardless of which alignment method is used beforehand. Our filtering therefore targets stylistic and semantic divergence among references that are already aligned to the source: a translation can be faithful yet very different in style.

\begin{figure}[h]
\centering
\includegraphics[width=0.98\linewidth]{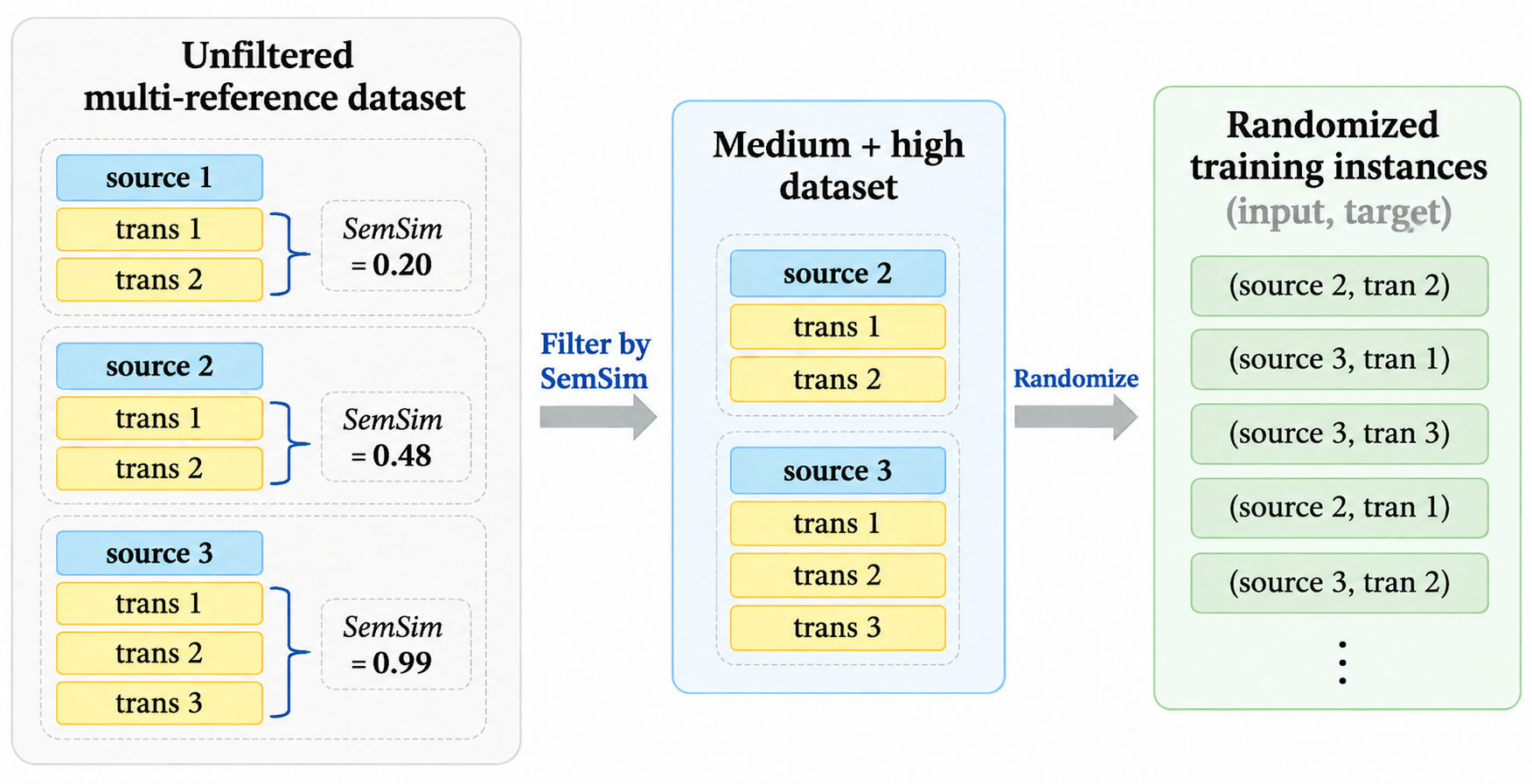}
\caption{The pipeline of creating the medium+high semantic similarity dataset from the unfiltered dataset. \semsim measures the semantic similarity among a source text's target translations.}
\label{fig:flow-chart}
\end{figure}

\begin{table*}[h]
\centering
\small
\resizebox{\textwidth}{!}{%
\begin{tabular}{lcc ccc ccc ccc}
\toprule
\multicolumn{12}{c}{\textbf{French--English (FR--EN)}} \\
\midrule
& \#src & \#instn
& \multicolumn{3}{c}{\textbf{BLEU (mean/std)}}
& \multicolumn{3}{c}{\textbf{COMET (mean/std)}}
& \multicolumn{3}{c}{\textbf{chrF++ (mean/std)}} \\
\cmidrule(lr){4-6} \cmidrule(lr){7-9} \cmidrule(lr){10-12}
\shortstack{\textbf{SemSim}\\\textbf{Level}}
& &
& mT5 & Gemma & OpusMT
& mT5 & Gemma & OpusMT
& mT5 & Gemma & OpusMT \\
\midrule
low    
& \orangebarcell{5016}{91647}  & \bluebarcell{11279}{91647}
& \ms{19.5}{3.69} & \ms{19.6}{3.47} & \ms{24}{0.5}
& \ms{0.63}{0.03} & \ms{0.64}{0.02} & \ms{0.65}{0}
& \ms{0.36}{0.05} & \ms{0.37}{0.02} & \ms{0.42}{0} \\

medium 
& \orangebarcell{4977}{91647}  & \bluebarcell{11279}{91647}
& \ms{34.7}{0.35} & \ms{25.3}{1.7} & \msb{30.6}{0.12}
& \ms{0.71}{0} & \ms{0.7}{0.01} & \msb{0.68}{0}
& \ms{0.50}{0.01} & \ms{0.44}{0.01} & \ms{0.47}{0.01} \\

high   
& \orangebarcell{4779}{91647}  & \bluebarcell{11279}{91647}
& \msb{35.0}{1.07} & \msb{30.6}{1.10} & \ms{30.4}{0.26}
& \msb{0.72}{0.01} & \msb{0.73}{0} & \msb{0.68}{0.01}
& \msb{0.52}{0.01} & \msb{0.50}{0.01} & \msb{0.49}{0} \\
\bottomrule
\end{tabular}

}
\end{table*}

\begin{table*}[h]
\centering
\small
\resizebox{\textwidth}{!}{%
  \begin{tabular}{lcc ccc ccc ccc}
\toprule
\multicolumn{12}{c}{\textbf{Russian--English (RU--EN)}} \\
\midrule
& \#src & \#instn 
& \multicolumn{3}{c}{\textbf{BLEU (mean/std)}}
& \multicolumn{3}{c}{\textbf{COMET (mean/std)}}
& \multicolumn{3}{c}{\textbf{chrF++ (mean/std)}} \\
\cmidrule(lr){4-6} \cmidrule(lr){7-9} \cmidrule(lr){10-12}
\shortstack{\textbf{SemSim}\\\textbf{Level}}
& &
& mT5 & Gemma & OpusMT
& mT5 & Gemma & OpusMT
& mT5 & Gemma & OpusMT \\
\midrule
low    
& \orangebarcell{1618}{91647} & \bluebarcell{4114}{91647}
& \ms{5.25}{0.21} & \ms{12.0}{0.38} & \ms{12.8}{1}
& \ms{0.54}{0} & \ms{0.58}{0.01} & \ms{0.57}{0.01}
& \ms{0.25}{0.01} & \ms{0.30}{0.01} & \ms{0.33}{0} \\

medium 
& \orangebarcell{1529}{91647}  & \bluebarcell{4114}{91647}
& \ms{19.2}{1.39} & \ms{18.3}{0.64} & \ms{18.6}{0.46}
& \ms{0.64}{0.01} & \ms{0.65}{0.01} & \ms{0.60}{0.01}
& \ms{0.38}{0} & \ms{0.39}{0.01}  & \ms{0.37}{0.01} \\

high   
& \orangebarcell{1874}{91647}  & \bluebarcell{4114}{91647}
& \msb{24.3}{0.36} & \msb{25.9}{1.56} & \msb{19.7}{0.49}
& \msb{0.65}{0} & \msb{0.68}{0} & \msb{0.61}{0.01}
& \msb{0.4}{0} & \msb{0.45}{0.01} & \msb{0.38}{0} \\
\bottomrule
\end{tabular}
}
\caption{Automatic metric results for mT5, Gemma, and OpusMT after fine-tuning on datasets with different semantic similarity (\semsim) levels: low, medium, and high. The top table uses the French-English language pair, and the bottom table uses Russian-English. To fairly evaluate the effect of \semsim on literary MT performance, the datasets contain the same number of training instances (\ninstances), and similar number of source sentences (\nsources). We report the mean and standard deviation across three different seeds. Medium and high \semsim produce comparable performance and both substantially outperform low \semsim.}
\label{tab:table_low_med_high_compare}
\end{table*}

\section{Human Evaluation} \label{section-human-eval}

To better understand and characterize the outputs from the different systems, we conducted a human evaluation using pairwise A/B comparisons. We recruited 48 translators from North America, Europe, and Africa, through Upwork\footnote{\url{https://www.upwork.com/}} and ProZ\footnote{ProZ is a platform for hiring freelance translators. \url{https://www.proz.com/}}.

The translators are native speakers of either French or English and fluent in the other language, and most have prior experience translating literary texts. All evaluations were completed using Label Studio\footnote{\url{https://labelstud.io}}. The translators were paid at the rate of 40 USD for approximately one hour of active annotation time, which is consistent with similar translation rates on Upwork and ProZ. 

In each evaluation task, translators were presented with 30 questions. Each question contained a source text in French and two English translations. They were then asked to choose which translation they preferred (A or B). If they considered the two translations to be equally good or bad, they could select ``tied''. Then the translators were asked to provide a brief justification for their choice. 

Detailed annotation instructions, along with an example evaluation question, are provided in Appendix~\ref{appendix-human-eval}. We discuss human evaluation results comparing different models in the following sections, alongside automatic evaluation results.

\section{Hypothesis 1: Different ranges of semantic similarity will lead to different performance.} \label{Section-finding-optimal}

We compute the \semsim score for the reference translations for each source text, and divide the source texts into three groups based on their \semsim scores: [-1, 0.45) is \emph{Low}, [0.45,0.85) is \emph{Medium}, and [0.85, 1] is \emph{High}. We summarize the linguistic characteristics of each range in Table~\ref{semantic_sim_table}. 

Each \semsim range is then used to fine-tune three different models described in Section~\ref{section-models-description} for literary machine translation. For a fair comparison, we keep the total number of training instances (\ninstances) the same, and total number of source texts (\nsources) as close as possible, and this isolates \semsim as the only changing variable. We run each model with three different seeds. Table~\ref{tab:table_low_med_high_compare} reports the mean and standard deviation for each automatic metric described in Appendix~\ref{section-auto-metrics}.

Across models, metrics, and language pairs, fine-tuning on high \semsim data on average gives the best performance, followed by  medium then low \semsim data. However, the performance gap between medium and high \semsim is much smaller than the gap between low and medium \semsim. 

From low to medium, mT5 shows the most gains, with BLEU increasing by 15.2 (+77.9\%), COMET by 0.08 (+12.7\%), and chrF++ by 0.14 (+38.9\%). Gemma and OpusMT also improve from low to medium, though with smaller relative gains.

\begin{table}[t]
\centering
\resizebox{\linewidth}{!}{%

\begin{tabular}{lccc}
\toprule
\textbf{Setting} & \textbf{Preferred A}& \textbf{Preferred B} & \textbf{Tied} \\
\midrule
high vs. low  &  75.8 & 10.8 &  13.3 \\
high vs. med  & 48.7 & 20.2 & 31.1 \\
med vs. low &  64.2 & 12.5  & 23.3 \\
\bottomrule
\end{tabular}
}
\caption{Human annotators' vote percentages from A/B testing of outputs produced by mT5-large fine-tuned on datasets with different \semsim levels: high, medium, and low. Annotators strongly prefer high and medium \semsim outputs over low \semsim outputs. }
\label{table_human_low_med_high}
\end{table}

\begin{figure}[t]
\centering

\begin{minipage}{0.48\linewidth}
    \centering
    \includegraphics[width=\linewidth]{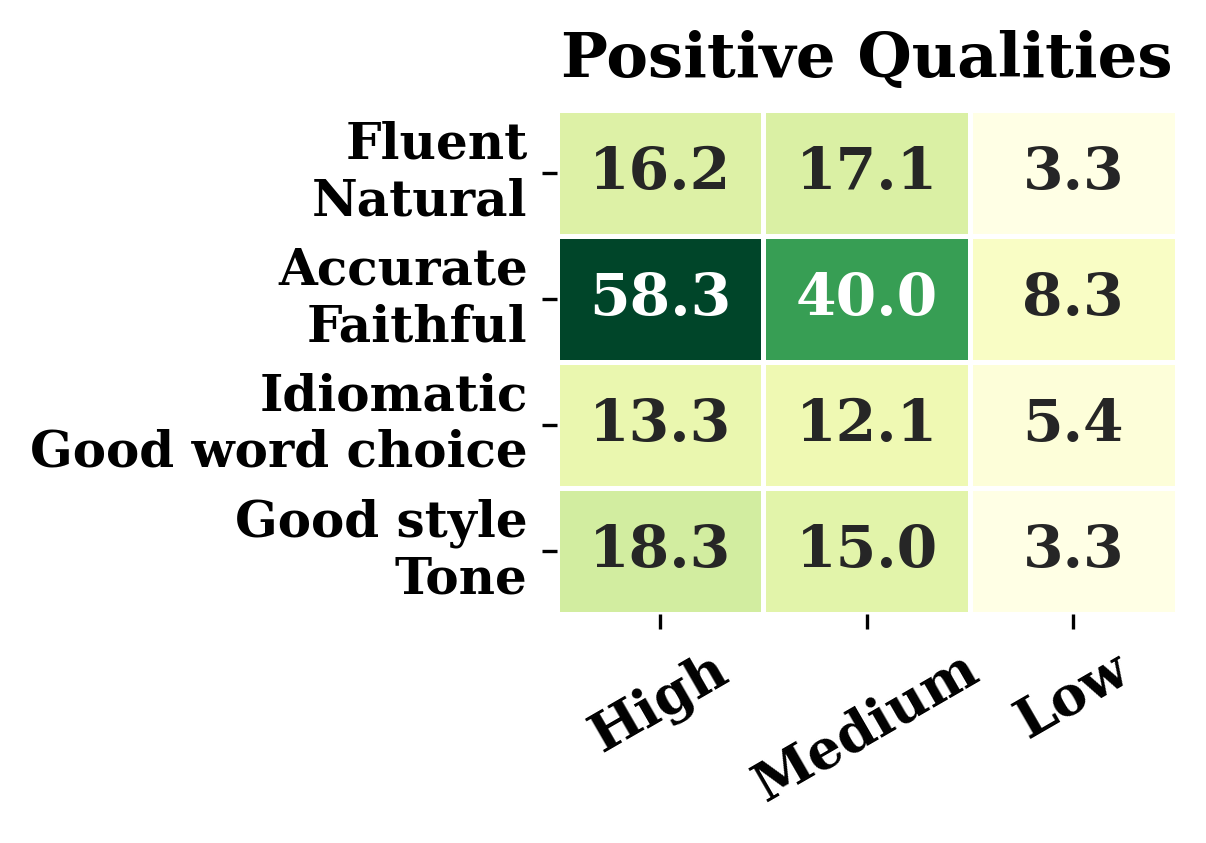}
\end{minipage}
\hfill
\begin{minipage}{0.48\linewidth}
    \centering
    \includegraphics[width=\linewidth]{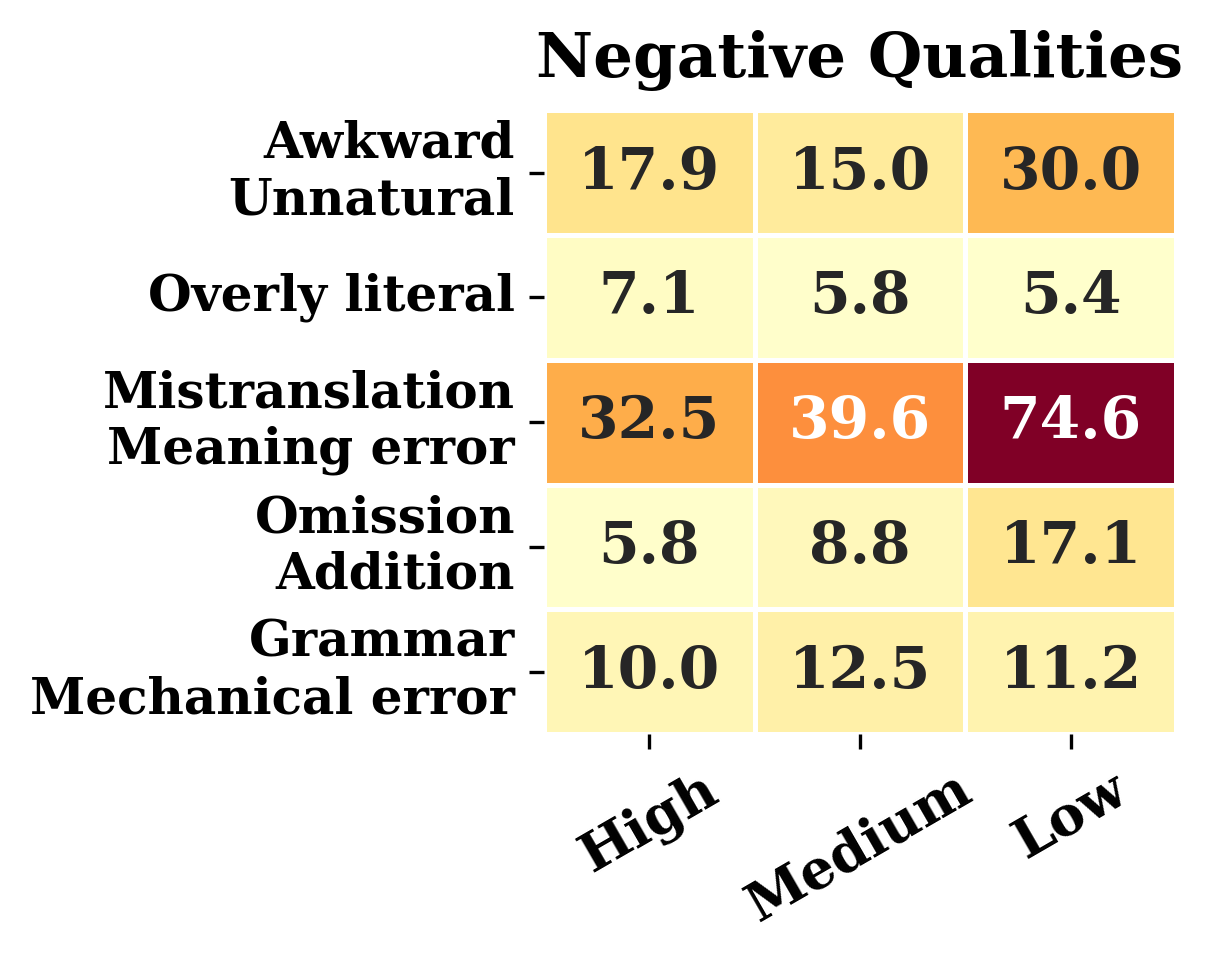}
\end{minipage}

\caption{Frequency (\%) of translation quality attributes mentioned in annotator comments. Each cell represents the percentage of comments where annotators explicitly noted a positive or negative quality attribute for a given system. High and medium \semsim data produce translations that are more fluent, accurate, idiomatic, and have better style and fewer errors than low \semsim.}
\label{fig:heatmap_HML}
\end{figure}

In contrast, the performance difference between medium and high is much smaller for mT5 and OpusMT: BLEU changes by less than 1 ($\pm 1\%$), COMET by at most 0.01 ($\leq 1.4\%$), and chrF++ by 0.02 ($\leq 4.3\%$). On the other hand, fine-tuning using Gemma, the performance difference between medium and high is only slightly less than low to medium: BLEU +5.3 (20.9\%), COMET +0.03 (4.3\%), and chrF++ increases by 0.06 (13.6\%).

These results suggest that medium and high \semsim data are substantially better for fine-tuning literary MT than low \semsim data. 

To assess whether these automatic metric trends align with human judgments, we hired 12 professional translators to conduct a pairwise comparison among mT5 outputs generated by fine-tuning on low, medium, and high \semsim data.

Source texts were randomly selected. We created four survey variants, with each variant assigned to three translators, so that each question was rated by three translators. No source text was repeated within a survey, across survey variants, or in any other experiment in this paper. 

\begin{table*}[t]
\centering
\small
\resizebox{\textwidth}{!}{%
\begin{tabular}{lcc ccc ccc ccc}
\toprule
\multicolumn{12}{c}{\textbf{French--English (FR--EN)}} \\
\midrule
& \#src & \#instn
& \multicolumn{3}{c}{\textbf{BLEU (mean/std)}}
& \multicolumn{3}{c}{\textbf{COMET (mean/std)}}
& \multicolumn{3}{c}{\textbf{chrF++ (mean/std)}} \\
\cmidrule(lr){4-6} \cmidrule(lr){7-9} \cmidrule(lr){10-12}
\textbf{SemSim Level}
& &
& mT5 & Gemma & OpusMT
& mT5 & Gemma & OpusMT
& mT5 & Gemma & OpusMT \\
\midrule

medium+low 
& \orangebarcell{30580}{91647}  & \bluebarcell{69575}{91647}
& \ms{40.6}{1.29} & \ms{27.2}{1.15} & \ms{32.7}{0.21}
& \ms{0.74}{0.01} & \ms{0.71}{0.01} & \ms{0.7}{0}
& \ms{0.53}{0.01} & \ms{0.46}{0.01} & \ms{0.5}{0} \\

medium+high 
& \orangebarcell{30595}{91647}& \bluebarcell{69575}{91647} 
& \ms{43.9}{0.50} & \msb{27.4}{1.0} & \ms{35.0}{0.21}
& \ms{0.74}{0} & \ms{0.71}{0.01} & \msb{0.71}{0.01}
& \ms{0.55}{0} & \ms{0.46}{0.02} & \msb{0.51}{0} \\

unfiltered 
& \orangebarcell{30580}{91647} & \bluebarcell{69575}{91647} 
& \ms{42.7}{0.40} & \msb{27.4}{1.15} & \ms{34.0}{0.67}
& \ms{0.74}{0} & \ms{0.71}{0} & \ms{0.70}{0.01}
& \ms{0.55}{0} & \ms{0.46}{0.01} & \msb{0.51}{0} \\
\midrule
unfiltered (all)
& \orangebarcell{39880}{91647}  & \bluebarcell{91647}{91647} 
& \ms{42.8}{1.53} & \ms{27.2}{1.44} & \ms{33.6}{0.25}
& \ms{0.74}{0.01} & \ms{0.71}{0.01} & \msb{0.71}{0}
& \ms{0.55}{0.01} & \ms{0.46}{0.01} & \msb{0.51}{0} \\

medium+high (all)
& \orangebarcell{ 34864 }{91647}    &  \bluebarcell{80368}{91647}
& \msb{44.3}{0.60} & \ms{26.8}{0.26} & \msb{35.2}{0.31}
& \msb{0.75}{0} & \ms{0.71}{0} & \msb{0.71}{0.01}
& \msb{0.56}{0.01} & \ms{0.46}{0} & \msb{0.51}{0.01} \\

\bottomrule
\end{tabular}

}
\end{table*}

\begin{table*}[h]
\centering
\small
\resizebox{\textwidth}{!}{%
  \begin{tabular}{lcc ccc ccc ccc}
\toprule
\multicolumn{12}{c}{\textbf{Russian--English (RU--EN)}} \\
\midrule
& \#src & \#instn 
& \multicolumn{3}{c}{\textbf{BLEU (mean/std)}}
& \multicolumn{3}{c}{\textbf{COMET (mean/std)}}
& \multicolumn{3}{c}{\textbf{chrF++ (mean/std)}} \\
\cmidrule(lr){4-6} \cmidrule(lr){7-9} \cmidrule(lr){10-12}
\shortstack{\textbf{SemSim}\\\textbf{Level}}
& &
& mT5 & Gemma & OpusMT
& mT5 & Gemma & OpusMT
& mT5 & Gemma & OpusMT \\
\midrule
medium+low 
& \orangebarcell{17530}{91647} & \bluebarcell{47641}{91647}
& \ms{29.7}{1.63} & \ms{22.8}{0.74} & \ms{24.9}{0.53}
& \ms{0.67}{0.01} & \ms{0.66}{0} & \ms{0.63}{0}
& \ms{0.45}{0.01} & \ms{0.41}{0} & \ms{0.42}{0} \\

medium+high 
&  \orangebarcell{17969 }{91647} & \bluebarcell{47641}{91647} 
& \ms{31.6}{0.64} & \ms{23.6}{1.46} & \ms{25.9}{0.06}
& \ms{0.67}{0.01} & \msb{0.67}{0.01} & \ms{0.64}{0.01}
& \ms{0.45}{0.01} & \ms{0.43}{0.01} & \ms{0.43}{0.01} \\

unfiltered 
&  \orangebarcell{18744}{91647} & \bluebarcell{47641}{91647}
& \ms{33.1}{0.37} & \msb{24.9}{0.76} & \ms{26.3}{0.36}
& \msb{0.68}{0} & \msb{0.67}{0} & \ms{0.64}{0}
& \msb{0.46}{0.01} & \msb{0.44}{0.01} & \ms{0.43}{0.01} \\
\midrule
unfiltered (all)
& \orangebarcell{26502}{91647}  & \bluebarcell{67106}{91647} 
& \msb{33.6}{0.49} & \ms{24.5}{0.8} & \ms{27.8}{0.32}
& \msb{0.68}{0} & \ms{0.66}{0.01} & \ms{0.64}{0.01}
& \msb{0.46}{0.01} & \ms{0.42}{0.01} & \msb{0.44}{0} \\

medium+high (all)
&  \orangebarcell{24884}{91647} & \bluebarcell{62992}{91647} 
& \ms{33.1}{0.96} & \ms{24.6}{0.96} & \msb{28}{0.1}
& \msb{0.68}{0.01} & \msb{0.67}{0} & \msb{0.65}{0.01}
& \msb{0.46}{0.01} & \ms{0.43}{0.01} & \msb{0.44}{0} \\

\bottomrule
\end{tabular}
}
\caption{The top section compares models fine-tuned on medium+low, medium+high, and unfiltered datasets, where all three datasets contain the same number of training instances and a similar number of source texts. The bottom section compares fine-tuning on the full unfiltered dataset with fine-tuning on the full medium+high \semsim subset. We report the mean and standard deviation across three different seeds. In both settings, medium+high \semsim gives comparable or better performance than unfiltered data.}
\label{tab:table_med_high_unfiltered_compare}
\end{table*}

Table~\ref{table_human_low_med_high} summarizes translators' votes as percentages in the pairwise comparison study. Translators strongly preferred  high and medium \semsim outputs over low \semsim outputs, with preference rates of 75.8\% vs. 10.8\%, and 64.2\% vs. 12.5\%, respectively. By contrast, the preference difference is smaller between high and medium \semsim , with translators preferring high over medium by 48.7\% vs. 20.2\%, and many more votes on ``tied'' (31.1\%).

To understand the linguistic characteristics of the outputs, we classify translators' comments on these pairwise comparisons using Claude Opus 4.6 API and visualize them in Figure~\ref{fig:heatmap_HML}. For each comment, we identify any positive and negative qualities of the translation mentioned, and we report how frequently the outputs from a system were mentioned with this quality. We provide the prompt for classifying their votes using Claude in Appendix~\ref{appendix-classify-translator-comment}.

Translators described high and medium \semsim outputs as comparably fluent, natural, idiomatic, with high \semsim as slightly better than medium \semsim in accuracy/faithfulness and style/tone, while low \semsim outputs were rarely associated with positive qualities.

Conversely, translators found low \semsim outputs to contain substantially more mistranslation,  omission and/or addition, and to be more awkward and unnatural. Many negative qualities were mentioned nearly twice as often for low \semsim than medium and high \semsim. Other negative qualities, such as overly literal and grammatical errors, occurred much less often and at similar rates as medium and high \semsim.

Overall, the human evaluation suggests that the automatic metric trends are consistent with human judgments. In particular, using medium and high \semsim data instead of low \semsim data improves perceived accuracy, style, word choice, and fluency, and substantially reduces mistranslation.

\section{Hypothesis 2: A filtered dataset will lead to better performance than an unfiltered dataset.} \label{section-combine-ranges}

Hypothesis~1 shows that both medium and high \semsim data produce substantially better automatic metric and human evaluation results than low \semsim data, and the performance between medium and high \semsim is comparable. We therefore investigate whether medium and high \semsim examples should be combined in practical fine-tuning scenarios. Prior work has also found that, when data quality is comparable, increasing the amount of training data can generally improve model performance \citep{madnani-etal-2007-using, madnani-etal-2008-multiple}.

Similar to Hypothesis~1, we fine-tune each model with three different random seeds and report the mean and standard deviation in Table~\ref{tab:table_med_high_unfiltered_compare}. We compare medium+high data against the unfiltered dataset under two settings: using all available examples, and using controlled subsets with the same number of sources and instances. Although unfiltered (all) contains 14\% more training instances than medium+high (all), medium+high (all) performs better across most models and metrics. In particular, mT5 improves from 42.8 to 44.3 BLEU, 0.74 to 0.75 COMET, and 0.55 to 0.56 chrF++; OpusMT improves from 33.6 to 35.2 BLEU while maintaining comparable COMET and chrF++. Gemma is the exception, where the two settings perform similarly.

\begin{table}[t]
\centering
\resizebox{\linewidth}{!}{%
\begin{tabular}{lcccc}
\toprule
\textbf{Setting} & \% &\textbf{BLEU}& \textbf{COMET}& \textbf{chrF++} \\
\midrule
FT w/ unfiltered & 37.8 & 45.3  & 0.78  & 0.61 \\
FT w/ med+high all & 36.1 & 45.7  & 0.78 & 0.61  \\
Tied & 26.1 &- &- &- \\ 
\bottomrule
\end{tabular}
}
\caption{Human annotators' vote percentages from A/B testing of outputs produced by mT5 fine-tuned on the full unfiltered and the full medium+high French-English datasets.}
\label{table_human_bestMT_vs_unfiltered}
\end{table}

We observe a similar pattern in the controlled comparison, where medium+low, medium+high, and unfiltered all have the same \nsources and \ninstances. To create medium+low and medium+high datasets, we keep \nsources and \ninstances fixed by using the full medium \semsim subset and sampling the same amounts of low and high \semsim data. Unfiltered is randomly sampled to match \nsources and \ninstances as well. Under this setting, medium+high performs best overall. Compared with medium+low, medium+high improves BLEU by 3.3 for mT5 and 2.3  for OpusMT, and 0.7 for Gemma, while also giving small improvements in chrF++. Compared with the size-matched unfiltered dataset, medium+high also gives higher BLEU for all three models, while COMET and chrF++ remain comparable or slightly better. These results suggest that adding low \semsim data does not provide the same benefit as adding high \semsim data, even when the total amount of training data is held constant.

\begin{figure}[t]
\centering

\begin{minipage}{0.48\linewidth}
    \centering
    \includegraphics[width=\linewidth]{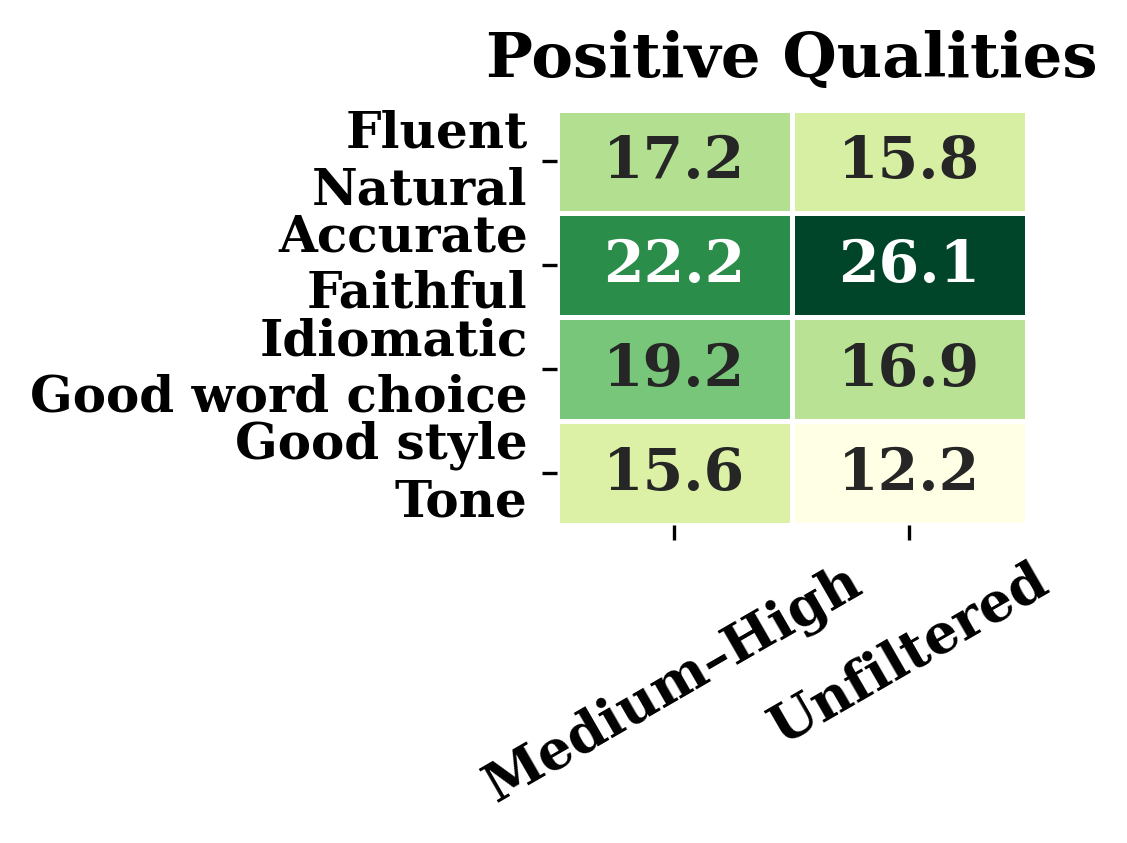}
\end{minipage}
\hfill
\begin{minipage}{0.48\linewidth}
    \centering
    \includegraphics[width=\linewidth]{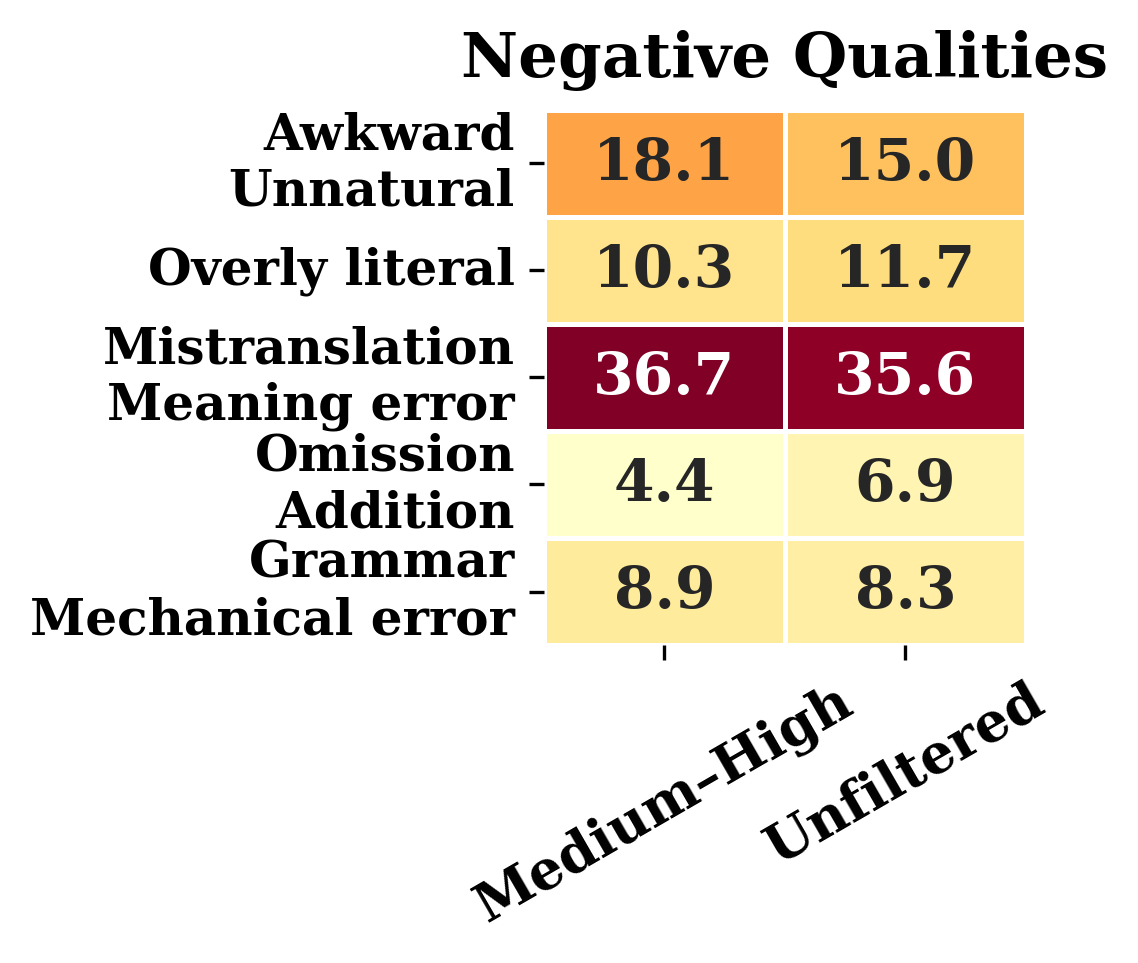}
\end{minipage}

\caption{Frequency (\%) of translation quality attributes mentioned in annotator comments. Annotators overall found the output from fine-tuning on the full unfiltered data and the full medium+high data to be comparable. }
\label{fig:heatmap_medium_high_unfiltered}
\end{figure}

\begin{table*}[t]
\centering
\small
\resizebox{\textwidth}{!}{%
  \begin{tabular}{lcc ccc ccc ccc}
\toprule
\multicolumn{12}{c}{\textbf{French--English (FR--EN)}} \\
\midrule
& \#src & \#instn 
& \multicolumn{3}{c}{\textbf{BLEU (mean/std)}}
& \multicolumn{3}{c}{\textbf{COMET (mean/std)}}
& \multicolumn{3}{c}{\textbf{chrF++ (mean/std)}} \\
\cmidrule(lr){4-6} \cmidrule(lr){7-9} \cmidrule(lr){10-12}
\shortstack{\textbf{Train Data}}
& &
& mT5 & Gemma & OpusMT
& mT5 & Gemma & OpusMT
& mT5 & Gemma & OpusMT \\
\midrule
human (single ref)   & \orangebarcell{34864}{91647}& \bluebarcell{34864}{91647} 
& \ms{41.4}{0.46} & \msb{28.8}{0.26} & \ms{33.5}{0.62}
& \ms{0.74}{0} & \msb{0.72}{0} & \ms{0.70}{0.01}
& \ms{0.54}{0} & \msb{0.47}{0.01} & \ms{0.50}{0.01} \\

human (multi ref) & \orangebarcell{34864}{91647} & \bluebarcell{80368}{91647} 
& \msb{44.3}{0.60} & \ms{26.8}{0.26} & \msb{35.2}{0.31}
& \msb{0.75}{0} & \ms{0.71}{0} & \msb{0.71}{0.01}
& \msb{0.56}{0.01} & \ms{0.46}{0} & \msb{0.51}{0.01} \\

synthetic (Gemma3) & \orangebarcell{34864}{91647} & \stackedbarcell{34864}{80368}{91647} 
& \ms{42.0}{0.1} & \ms{24.3}{0.32} & \ms{35.1}{0.31}
& \ms{0.74}{0} & \ms{0.7}{0} & \ms{0.7}{0}
& \ms{0.55}{0} & \ms{0.43}{0.01} & \msb{0.51}{0} \\

synthetic (Mistral) & \orangebarcell{34864}{91647}& \stackedbarcell{34864}{80368}{91647} 
& \ms{40.5}{0.4} & \ms{22.9}{0.89} & \ms{33.2}{0.78}
& \ms{0.74}{0} & \ms{0.7}{0} & \msb{0.71}{0}
& \ms{0.54}{0.01} & \ms{0.43}{0.01} & \msb{0.51}{0} \\

synthetic (Qwen3) & \orangebarcell{34864}{91647} & \stackedbarcell{34864}{80368}{91647} 
& \ms{40.2}{2.35} & \ms{26.0}{0.92} & \ms{34.3}{0.35}
& \ms{0.72}{0.01} & \ms{0.71}{0.01} & \ms{0.7}{0}
& \ms{0.50}{0.01} & \ms{0.44}{0.01} & \msb{0.51}{0} \\

\bottomrule
\end{tabular}
}
\end{table*}

\begin{table*}[h]
\centering
\small
\resizebox{\textwidth}{!}{%
  \begin{tabular}{lcc ccc ccc ccc}
\toprule
\multicolumn{12}{c}{\textbf{Russian--English (RU--EN)}} \\
\midrule
& \#src & \#instn 
& \multicolumn{3}{c}{\textbf{BLEU (mean/std)}}
& \multicolumn{3}{c}{\textbf{COMET (mean/std)}}
& \multicolumn{3}{c}{\textbf{chrF++ (mean/std)}} \\
\cmidrule(lr){4-6} \cmidrule(lr){7-9} \cmidrule(lr){10-12}
\shortstack{\textbf{Train}\\\textbf{Data}}
& &
& mT5 & Gemma & OpusMT
& mT5 & Gemma & OpusMT
& mT5 & Gemma & OpusMT \\
\midrule
human (single ref)   
&\orangebarcell{24884}{91647}  & \bluebarcell{24884}{91647}
& \ms{31.9}{0.66} & \ms{23.6}{1.06} & \ms{25.6}{0.98}
& \msb{0.68}{0.01} & \msb{0.67}{0.01} & \ms{0.64}{0}
& \ms{0.45}{0.01} & \ms{0.34}{0.01} & \ms{0.43}{0} \\

human (multi-ref) 
& \orangebarcell{24884}{91647} & \bluebarcell{62992}{91647}
& \msb{33.1}{0.96} & \msb{24.6}{0.96} & \msb{28}{0.1}
& \msb{0.68}{0.01} & \msb{0.67}{0} & \msb{0.65}{0.01}
& \msb{0.46}{0.01} & \msb{0.43}{0.01} & \msb{0.44}{0} \\

synthetic (Gemma3) 
&  \orangebarcell{24884}{91647}& \stackedbarcell{24884}{62992}{91647}  
& \ms{18.9}{1.53} & \ms{11.9}{0.69} & \ms{25.2}{0.12}
& \ms{0.63}{0} & \ms{0.61}{0.01} & \ms{0.63}{0.01}
& \ms{0.39}{0.01} & \ms{0.32}{0.01} & \ms{0.41}{0.01} \\

synthetic (Mistral) 
& \orangebarcell{24884}{91647} & \stackedbarcell{24884}{62992}{91647}
& \ms{23.2}{1.08} & \ms{13}{1.24} & \ms{25.1}{0.25}
& \ms{0.65}{0.01} & \ms{0.62}{0} & \ms{0.63}{0.01}
& \ms{0.40}{0.01} & \ms{0.34}{0.01} & \ms{0.42}{0.01} \\

synthetic (Qwen3) 
& \orangebarcell{24884}{91647} & \stackedbarcell{24884}{62992}{91647}
& \ms{20.7}{1.63} & \ms{14.1}{1.13} & \ms{25.3}{0.1}
& \ms{0.64}{0.01} & \ms{0.63}{0} & \ms{0.63}{0}
& \ms{0.39}{0} & \ms{0.34}{0.01} & \ms{0.41}{0.01} \\

\bottomrule
\end{tabular}
}
\caption{Performance comparison between human-only and synthetic-augmented training data. ``Human multi ref'' denotes training on the full medium and high \semsim data with multiple human references, whereas ``Human single ref'' uses the same source texts with one randomly selected human reference per source. ``Synthetic (Gemma3)'' denotes training data created by using instruction-tuned Gemma3 to create additional target texts. We report the mean and standard deviation across three different seeds. Synthetic-augmented training data underperform human multi-reference data, but synthetic references generated by a strong instruction-tuned model such as Gemma3 can improve MT performance over a single-reference dataset when fine-tuning using mT5 and OpusMT.}
\label{table_human_synthetic_naive}
\end{table*}

Translators also judged the two systems to be very close, preferring unfiltered in 37.8\%, medium+high in 36.1\%, with 26.1\% marked as ties. Notably, the tie rate is large relative to the preference rates for either system, suggesting that translators often found the two outputs comparable.

The analysis on positive and negative qualities also shows a similarly mixed pattern: medium+high has a slight advantage in fluency, word choice/idiomatic, and style/tone, while unfiltered is described slightly more often as accurate. Interestingly, the positive and negative comments are almost contradictory: medium+high is more fluent/natural but also more awkward/unnatural, unfiltered is more accurate/faithful but also has more omission/addition.

Many translators also commented on the limitations of the setup. Translators noted that some translations are hard to decide which is better without even longer context such as the preceding or following sentence around a paragraph. This was especially true for abrupt paragraphs such as quotations, which are common in literary text. In such cases, some translators preferred a more literal translation to be safe, while others favored a more natural and idiomatic interpretation of the original text. A natural transition at the start of a paragraph can, in isolation, look like a mistranslation without more context, especially if a translator took creative liberties with the translation (artistic license). Translators also noted that character names were sometimes translated by both systems. While some translators considered this a serious flaw, others overlooked it.

Overall, the automatic metrics and human evaluation suggest that medium+high data is at least competitive with the unfiltered dataset, despite the unfiltered dataset containing 14\% more training examples. The controlled comparison further shows that replacing high \semsim examples with low \semsim examples hurts performance, supporting the decision to exclude low \semsim data.

\section{Hypothesis 3: A synthetic-augmented multi-reference dataset will perform comparably to an expert-translated dataset.} \label{Section-synthetic}

We investigate whether augmenting a single-reference dataset with synthetic references can achieve  performance comparable to that of expert-translated multi-reference datasets.

Specifically, for each source text in medium+high, we randomly select one human translation and generate paraphrases of this translation using open-weight instruction-tuned LLMs. For all the multi-reference datasets, we keep the number of target texts per source text the same. We provide model details and the prompts we used for zero-shot paraphrase generation in Appendix~\ref{appendix-synthetic-model-prompts}. We compare using these synthetic-augmented data against our overall best-performing model, which is fine-tuned on the full medium and high \semsim data.

We find that synthetic-augmented datasets provide minimal improvements over the single-reference dataset, or in some cases, even degrade the performance depending on the model used to generate the synthetic references (Table~\ref{table_human_synthetic_naive}). The best model for generating synthetic references, Gemma3, increases BLEU by 0.6 when fine-tuning with mT5, and 1.6 with OpusMT, and improves COMET by 0.01 for both mT5 and OpusMT. However, none of the synthetic datasets improve performance when fine-tuned using Gemma. 

\begin{table}[t]
\centering
\resizebox{\linewidth}{!}{%
\begin{tabular}{lcccc}
\toprule
\textbf{Setting} & \textbf{\%} & \textbf{BLEU}& \textbf{COMET}& \textbf{chrF++}\\
\midrule
\textbf{FT w/ human} &  \textbf{50.4} &45.1 & 0.79 & 0.60\\
FT w/ human + synthetic & 32.1 & 41.0 & 0.75&0.56 \\
Tied & 17.5 & - & - & - \\
\bottomrule
\end{tabular}
}
\caption{Human annotators' vote percentages from A/B testing of outputs produced by mT5 fine-tuned on the full medium and high \semsim data and the synthetically augmented data. Synthetic data is generated using instruction-tuned Gemma3. Annotators preferred outputs of the MT system fine-tuned on the human data instead of synthetically augmented data by 18\%.}
\label{table_human_bestMT_vs_syn}
\end{table}

\begin{figure}[t]
\centering

\begin{minipage}{0.48\linewidth}
    \centering
    \includegraphics[width=\linewidth]{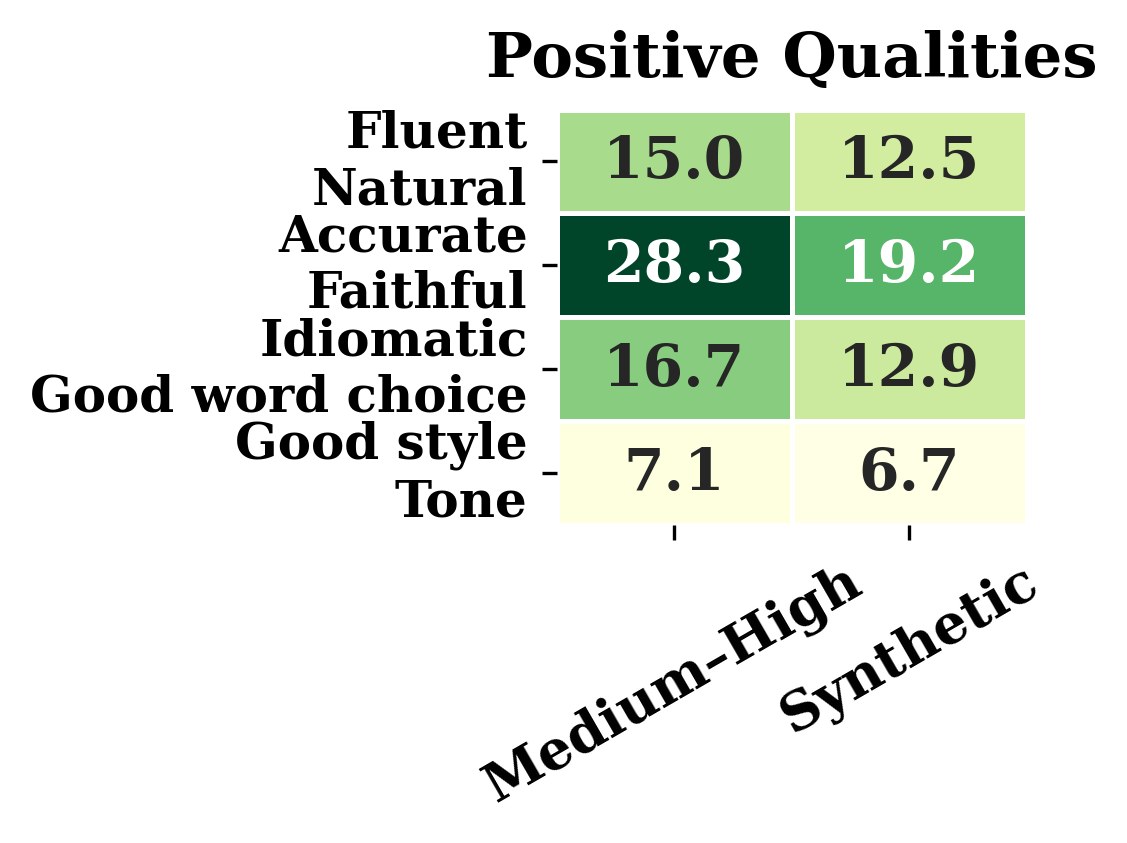}
\end{minipage}
\hfill
\begin{minipage}{0.48\linewidth}
    \centering
    \includegraphics[width=\linewidth]{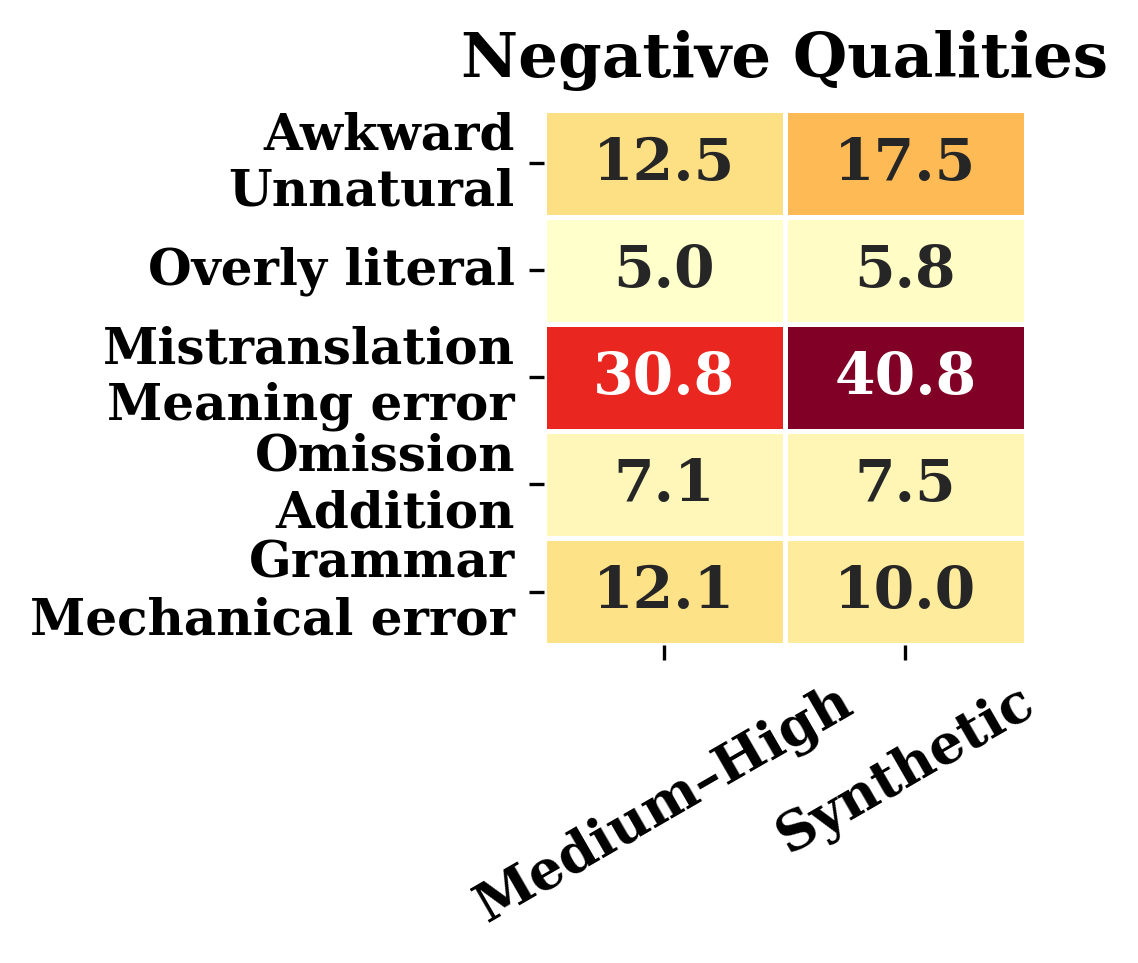}
\end{minipage}

\caption{Frequency (\%) of translation quality attributes mentioned in annotator comments. Translation outputs from model trained on human data (the full medium and high \semsim data) were found to be more accurate, idiomatic, and have less mistranslation than model trained on synthetically augmented data. }
\label{fig:heatmap_medium_high_synthetic}
\end{figure}

Moreover, none of the synthetic datasets outperform the  medium+high multi-reference dataset under any model or metric. We further conduct a human evaluation comparing medium+high with the best synthetic-augmented dataset using mT5.

In Figure~\ref{fig-syn-simp-violin-ecdf} in the appendix, we also compare the distribution of the pairwise \spavgfull scores between the ground-truth human translation and each generated synthetic translation. Compared to the \spavgfull scores between two human translations, the score between a human and a synthetic translation is on average lower across all models. The middle 50\% of the distribution shifts down, as does the median. This means the \semsim of synthetically augmented data is lower overall.

Translators preferred translations from model fine-tuned on human data over synthetic-augmented data, with preference votes of 50.4\% vs. 32.1\%, and 17.5\% tied.

Qualitative analysis on translators' comments further shows that expert-translated data leads to more accurate, idiomatic, and fluent translations. 

Lastly, the human evaluation also includes comparisons between ground-truth expert translations and outputs from our best fine-tuned model.

Unsurprisingly, translators preferred human translations over machine translated text (55.4\% vs. 31.3\%). Human translations were judged to be substantially more fluent/natural (27.1\% vs. 8.8\%), more idiomatic (27.1\% vs. 9.6\%), and have much less mistranslation (12.1\% vs. 34.6\%).

Interestingly, translators found human translations more often involved omissions and additions (18.8\% vs 5.8\%), echoing the concerns raised in Hypothesis 2, where  translators noted that some comparisons are difficult to judge without additional context. For example, if the name ``Christine'' appears in a previous paragraph, a human translator may use ``she'' in the following paragraph to refer to Christine, even if the original French text repeats ``Christine'' in that paragraph. In contrast, machine translated text is more likely to retain ``Christine'' in order to match the French source text.

\begin{table}[t]
\centering
\small
\begin{tabular}{lc}
\toprule
\textbf{Setting} & \textbf{Percentage} \\
\midrule
human GT  &  55.4\\
FT w/ med+high all  & 31.3 \\
Tied & 13.3 \\
\bottomrule
\end{tabular}

\caption{Human annotators' vote percentages from A/B testing of outputs produced by mT5 fine-tuned on the full medium and high \semsim data (denoted as ``FT w/ med+high all'') and ground-truth human expert translations (denoted as ``human GT'').}
\label{table_human_bestMT_vs_human}
\end{table}

\begin{figure}[t]
\centering

\begin{minipage}{0.48\linewidth}
    \centering
    \includegraphics[width=\linewidth]{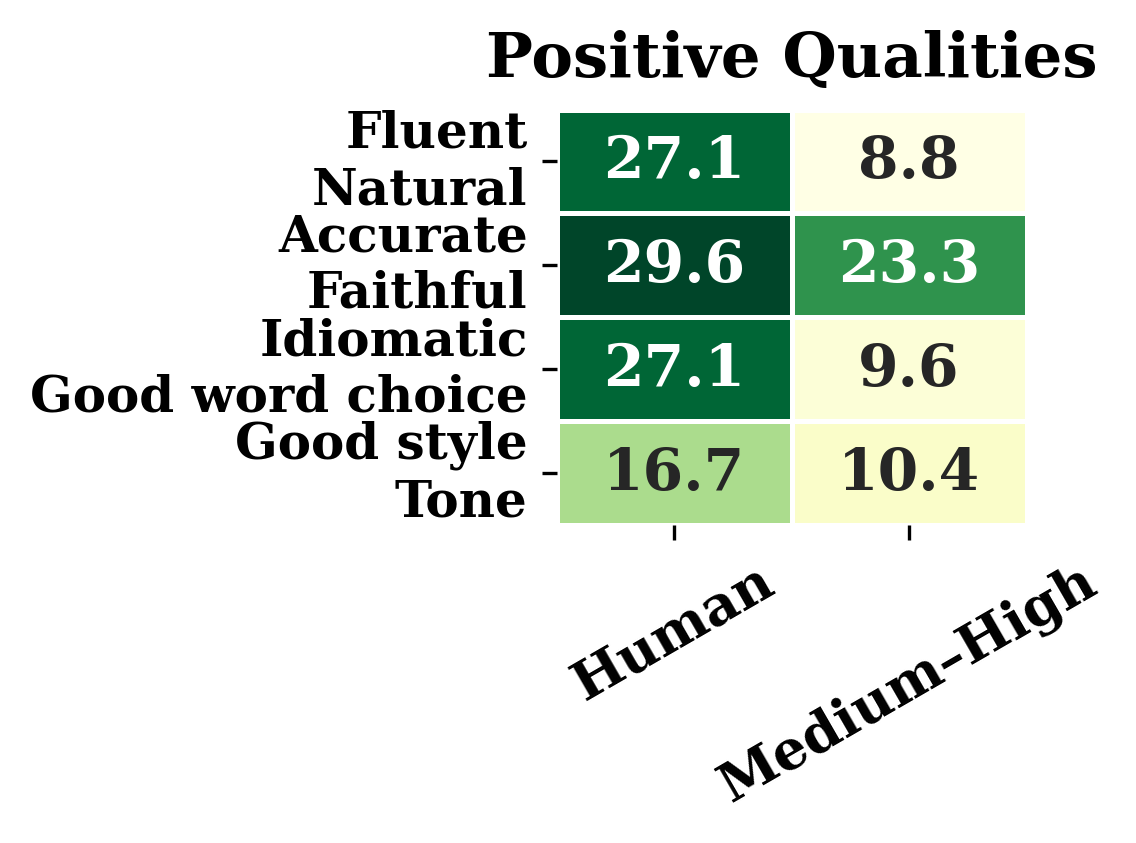}
\end{minipage}
\hfill
\begin{minipage}{0.48\linewidth}
    \centering
    \includegraphics[width=\linewidth]{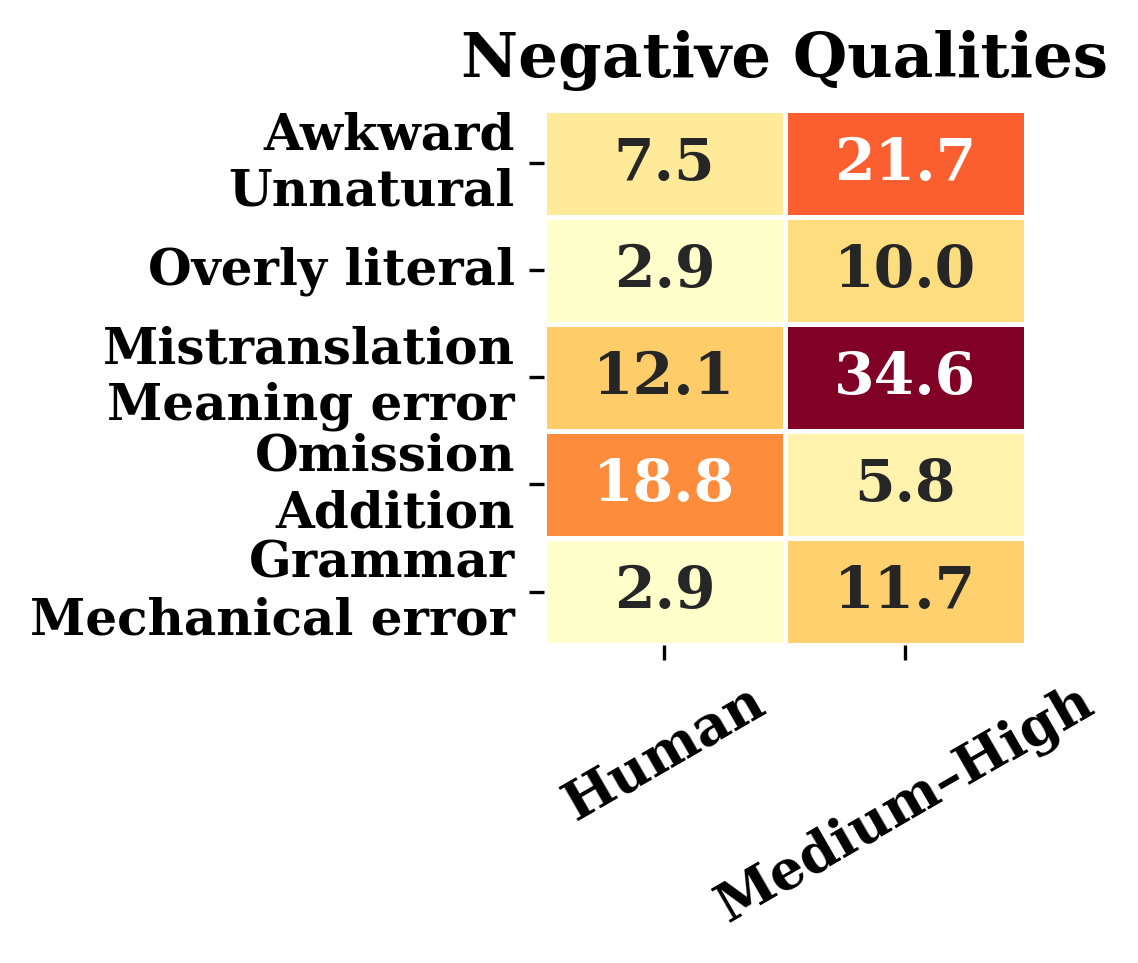}
\end{minipage}

\caption{Ground-truth human expert translation were more fluent, idiomatic, and have better style, though more prone to omission and addition than our best MT system output.}
\label{fig:heatmap_human_medium_high}
\end{figure}

\section{Conclusion} \label{section-conclusion}

Most MT fine-tuning work treats references as interchangeable ground truth, using a single reference per source or weighting all references equally, without considering whether the references agree with one another. In contrast, we use the semantic similarity among multiple references as a filtering signal---an idea motivated by the comparative-literature practice of reading multiple translations. To our knowledge, this is the first work to do so for literary MT.

We propose a framework for filtering a multi-reference dataset for literary MT, a setting where multiple translations of a source text often exist. Results from automatic metrics on both FR-EN and RU-EN show that high and medium semantic similarity data is better for fine-tuning MT systems than low semantic similarity data. Human evaluation confirms this: translators overwhelmingly prefer systems fine-tuned on high \semsim over low, and medium over low. When commenting on the outputs, translators rarely associate any positive qualities with low \semsim, while their assessments of high and medium are more comparable, with high \semsim holding a slight edge. Importantly, references with low semantic similarity among them are not merely misaligned or erroneous; they are also cases where translators diverge substantially in their interpretations.

We also find that while using more data generally improves performance, unfiltered data does not always help. Medium+high data can give comparable and at times better performance, all with less data. Human translators also found outputs from medium+high data more idiomatic and better in style, though overall performance is comparable to using unfiltered data. Translators also noted limitations of the setup that made comparison difficult: paragraphs were presented without preceding or following context, and translators differed in their own preferences, favoring more literal versus more natural, idiomatic interpretations.

We are not aware of any prior work that compares fine-tuning using human versus synthetic-augmented data for literary MT, particularly with human evaluation that examines the language of the output translations themselves. Comparing synthetic against human translations, we find that synthetic data, unlike human data, does not reliably improve MT. Performance depends heavily on the LLM used to paraphrase the ground-truth translation. For French-English, of the three LLMs we tested, only Gemma3 improved MT performance compared to using a single reference, and the improvement was small. For Russian-English, performance worsened, suggesting that paraphrase errors and deviation from the original translation act as noise. In other words, augmenting a dataset by zero-shot LLM paraphrasing of references can hurt MT performance. Interestingly, the best paraphrasing LLM also had, on average, a lower \semsim, while remaining mostly in the medium \semsim range (Figure~\ref{fig-syn-simp-violin-ecdf}). 

This preference for human data is echoed in human evaluation: comparing outputs from models fine-tuned on human versus synthetic augmented data, translators preferred those trained on human data much more, finding them more accurate, better in word choice, and far less prone to mistranslation.

As a sanity check, we also compared the ground-truth translations with our best fine-tuned model (mT5 fine-tuned on all medium+high \semsim data), and translators preferred the ground truth at a rate similar to the earlier comparisons. While the ground-truth translations were most frequently credited with positive qualities, they were also, surprisingly, mentioned far more often for omission and addition. Considering how paragraphs flow, this makes sense: good flow often means being less literal, with translators moving content into the preceding or following paragraph.

Finally, we tested zero-shot translation (Appendix~\ref{section-zero-shot}). Compared to our best fine-tuned model (mT5 fine-tuned on all medium+high \semsim data), our model outperformed all open-weight models we tested. Among commercial models, two of three outperformed ours, one only marginally (44.8 vs. 44.3 in BLEU).

Meaningful human evaluation requires a sufficient amount of data, which limits us to high-resource language pairs, since low-quality output will also result in noisy human evaluation when errors are frequent and of many kinds. We hope our findings provide guidance for literary translation on both high- and low-resource language pairs: prioritize human translations, and if using synthetic LLM-based paraphrasing---even into a high-resource target language like English---choose the LLM carefully and inspect the paraphrased output, or performance may suffer.

Future work can compare LLM zero-shot paraphrasing with more controlled generation methods, such as pivot-based or neural approaches, especially for low-resource languages.

\section*{Limitations}
The following limitations of our paper are also directions for future work.

\begin{enumerate}
    \item We create additional synthetic references only using LLMs in this paper. Future work can explore different methods for generating synthetic references and compare their performance. More controlled generation could potentially prevent the overall semantic similarity shift demonstrated in Figure~\ref{fig-syn-simp-violin-ecdf}. 
        
    \item While we would have liked to work on low-resource languages in this paper, the premise of the paper relies on multi-reference data, which generally exists only for high-resource language pairs. Additionally, conducting human evaluation on low-quality translations in low-resource setting would be challenging. If the outputs are not adequate, it is difficult to obtain meaningful feedback from translators. However, we believe our findings can benefit future work on low-resource settings. Specifically, it is worth investigating whether filtering by semantic similarity between synthetic references can help improve low-resource MT.
    
    \item Due to our computational constraints, we could not experiment with LLMs with even larger numbers of parameters. For example, we could only fine-tune Gemma3 with 12 billion parameters using QLoRA, and its performance was, for the most part, worse than that of mT5-Large. 
        
    \item We would also like to extend the context window by training on longer data. As discussed in Hypothesis 2, translators mentioned that without the preceding and following paragraphs, it is sometimes difficult to judge which translation is better or whether there is a mistranslation. However, due to computational constraints and the nature of the data in Par3, we are limited to paragraph-level data, which is still better than sentence-level data. Ideally, book-chapter level translation would be a great next step.

\end{enumerate}

\section*{Acknowledgments}
We thank Huda Khayrallah and Brian Thompson for their early feedback on the paper.  Si Wu and David Smith were supported in part by the Schmidt Sciences-funded project ``Beyond Translation: Opening up the Human Record.''

\newpage

\bibliography{anthology,main_paper}

@inproceedings{besacier-schwartz-2015-automated,
    title = "Automated Translation of a Literary Work: A Pilot Study",
    author = "Besacier, Laurent  and
      Schwartz, Lane",
    booktitle = "Proceedings of the Fourth Workshop on Computational Linguistics for Literature",
    month = jun,
    year = "2015",
    address = "Denver, Colorado, USA",
    publisher = "Association for Computational Linguistics",
    url = "https://aclanthology.org/W15-0713",
    doi = "10.3115/v1/W15-0713",
    pages = "114--122",
}

@inproceedings{briakou-carpuat-2021-beyond,
    title = "Beyond Noise: Mitigating the Impact of Fine-grained Semantic Divergences on Neural Machine Translation",
    author = "Briakou, Eleftheria  and
      Carpuat, Marine",
    booktitle = "Proceedings of the 59th Annual Meeting of the Association for Computational Linguistics and the 11th International Joint Conference on Natural Language Processing (Volume 1: Long Papers)",
    month = aug,
    year = "2021",
    address = "Online",
    publisher = "Association for Computational Linguistics",
    url = "https://aclanthology.org/2021.acl-long.562",
    doi = "10.18653/v1/2021.acl-long.562",
    pages = "7236--7249",
}

@inproceedings{callison-burch-etal-2006-improved,
    title = "Improved Statistical Machine Translation Using Paraphrases",
    author = "Callison-Burch, Chris  and
      Koehn, Philipp  and
      Osborne, Miles",
    booktitle = "Proceedings of the Human Language Technology Conference of the {NAACL}, Main Conference",
    month = jun,
    year = "2006",
    address = "New York City, USA",
    publisher = "Association for Computational Linguistics",
    url = "https://aclanthology.org/N06-1003",
    pages = "17--24",
}

@inproceedings{carpuat-etal-2017-detecting,
    title = "Detecting Cross-Lingual Semantic Divergence for Neural Machine Translation",
    author = "Carpuat, Marine  and
      Vyas, Yogarshi  and
      Niu, Xing",
    booktitle = "Proceedings of the First Workshop on Neural Machine Translation",
    month = aug,
    year = "2017",
    address = "Vancouver",
    publisher = "Association for Computational Linguistics",
    url = "https://aclanthology.org/W17-3209",
    doi = "10.18653/v1/W17-3209",
    pages = "69--79",
}

@inproceedings{conneau-etal-2020-unsupervised,
    title = "Unsupervised Cross-lingual Representation Learning at Scale",
    author = "Conneau, Alexis  and
      Khandelwal, Kartikay  and
      Goyal, Naman  and
      Chaudhary, Vishrav  and
      Wenzek, Guillaume  and
      Guzm{\'a}n, Francisco  and
      Grave, Edouard  and
      Ott, Myle  and
      Zettlemoyer, Luke  and
      Stoyanov, Veselin",
    booktitle = "Proceedings of the 58th Annual Meeting of the Association for Computational Linguistics",
    month = jul,
    year = "2020",
    address = "Online",
    publisher = "Association for Computational Linguistics",
    url = "https://aclanthology.org/2020.acl-main.747",
    doi = "10.18653/v1/2020.acl-main.747",
    pages = "8440--8451",
}

@inproceedings{edunov-etal-2018-understanding,
    title = "Understanding Back-Translation at Scale",
    author = "Edunov, Sergey  and
      Ott, Myle  and
      Auli, Michael  and
      Grangier, David",
    booktitle = "Proceedings of the 2018 Conference on Empirical Methods in Natural Language Processing",
    month = oct # "-" # nov,
    year = "2018",
    address = "Brussels, Belgium",
    publisher = "Association for Computational Linguistics",
    url = "https://aclanthology.org/D18-1045",
    doi = "10.18653/v1/D18-1045",
    pages = "489--500",
}

@inproceedings{junczys-dowmunt-etal-2018-marian,
    title = "{M}arian: Fast Neural Machine Translation in {C}++",
    author = "Junczys-Dowmunt, Marcin  and
      Grundkiewicz, Roman  and
      Dwojak, Tomasz  and
      Hoang, Hieu  and
      Heafield, Kenneth  and
      Neckermann, Tom  and
      Seide, Frank  and
      Germann, Ulrich  and
      Aji, Alham Fikri  and
      Bogoychev, Nikolay  and
      Martins, Andr{\'e} F. T.  and
      Birch, Alexandra",
    booktitle = "Proceedings of {ACL} 2018, System Demonstrations",
    month = jul,
    year = "2018",
    address = "Melbourne, Australia",
    publisher = "Association for Computational Linguistics",
    url = "https://aclanthology.org/P18-4020",
    doi = "10.18653/v1/P18-4020",
    pages = "116--121",
}

@inproceedings{khayrallah-etal-2020-simulated,
    title = "Simulated multiple reference training improves low-resource machine translation",
    author = "Khayrallah, Huda  and
      Thompson, Brian  and
      Post, Matt  and
      Koehn, Philipp",
    booktitle = "Proceedings of the 2020 Conference on Empirical Methods in Natural Language Processing (EMNLP)",
    month = nov,
    year = "2020",
    address = "Online",
    publisher = "Association for Computational Linguistics",
    url = "https://aclanthology.org/2020.emnlp-main.7",
    doi = "10.18653/v1/2020.emnlp-main.7",
    pages = "82--89",
}

@inproceedings{khayrallah-koehn-2018-impact,
    title = "On the Impact of Various Types of Noise on Neural Machine Translation",
    author = "Khayrallah, Huda  and
      Koehn, Philipp",
    booktitle = "Proceedings of the 2nd Workshop on Neural Machine Translation and Generation",
    month = jul,
    year = "2018",
    address = "Melbourne, Australia",
    publisher = "Association for Computational Linguistics",
    url = "https://aclanthology.org/W18-2709",
    doi = "10.18653/v1/W18-2709",
    pages = "74--83",
}

@inproceedings{madnani-etal-2007-using,
    title = "Using Paraphrases for Parameter Tuning in Statistical Machine Translation",
    author = "Madnani, Nitin  and
      Fazil Ayan, Necip  and
      Resnik, Philip  and
      Dorr, Bonnie",
    booktitle = "Proceedings of the Second Workshop on Statistical Machine Translation",
    month = jun,
    year = "2007",
    address = "Prague, Czech Republic",
    publisher = "Association for Computational Linguistics",
    url = "https://aclanthology.org/W07-0716",
    pages = "120--127",
}

@inproceedings{madnani-etal-2008-multiple,
    title = "Are Multiple Reference Translations Necessary? Investigating the Value of Paraphrased Reference Translations in Parameter Optimization",
    author = "Madnani, Nitin  and
      Resnik, Philip  and
      Dorr, Bonnie J.  and
      Schwartz, Richard",
    booktitle = "Proceedings of the 8th Conference of the Association for Machine Translation in the Americas: Research Papers",
    month = oct # " 21-25",
    year = "2008",
    address = "Waikiki, USA",
    publisher = "Association for Machine Translation in the Americas",
    url = "https://aclanthology.org/2008.amta-papers.13",
    pages = "143--152",
}

@inproceedings{papineni-etal-2002-bleu,
    title = "{B}leu: a Method for Automatic Evaluation of Machine Translation",
    author = "Papineni, Kishore  and
      Roukos, Salim  and
      Ward, Todd  and
      Zhu, Wei-Jing",
    booktitle = "Proceedings of the 40th Annual Meeting of the Association for Computational Linguistics",
    month = jul,
    year = "2002",
    address = "Philadelphia, Pennsylvania, USA",
    publisher = "Association for Computational Linguistics",
    url = "https://aclanthology.org/P02-1040",
    doi = "10.3115/1073083.1073135",
    pages = "311--318",
}

@inproceedings{popovic-2017-chrf,
    title = "chr{F}++: words helping character n-grams",
    author = "Popovi{\'c}, Maja",
    booktitle = "Proceedings of the Second Conference on Machine Translation",
    month = sep,
    year = "2017",
    address = "Copenhagen, Denmark",
    publisher = "Association for Computational Linguistics",
    url = "https://aclanthology.org/W17-4770",
    doi = "10.18653/v1/W17-4770",
    pages = "612--618",
}

@inproceedings{rei-etal-2020-comet,
    title = "{COMET}: A Neural Framework for {MT} Evaluation",
    author = "Rei, Ricardo  and
      Stewart, Craig  and
      Farinha, Ana C  and
      Lavie, Alon",
    booktitle = "Proceedings of the 2020 Conference on Empirical Methods in Natural Language Processing (EMNLP)",
    month = nov,
    year = "2020",
    address = "Online",
    publisher = "Association for Computational Linguistics",
    url = "https://aclanthology.org/2020.emnlp-main.213",
    doi = "10.18653/v1/2020.emnlp-main.213",
    pages = "2685--2702",
}

@inproceedings{sennrich-etal-2016-improving,
    title = "Improving Neural Machine Translation Models with Monolingual Data",
    author = "Sennrich, Rico  and
      Haddow, Barry  and
      Birch, Alexandra",
    booktitle = "Proceedings of the 54th Annual Meeting of the Association for Computational Linguistics (Volume 1: Long Papers)",
    month = aug,
    year = "2016",
    address = "Berlin, Germany",
    publisher = "Association for Computational Linguistics",
    url = "https://aclanthology.org/P16-1009",
    doi = "10.18653/v1/P16-1009",
    pages = "86--96",
}

@inproceedings{shen-etal-2016-minimum,
    title = "Minimum Risk Training for Neural Machine Translation",
    author = "Shen, Shiqi  and
      Cheng, Yong  and
      He, Zhongjun  and
      He, Wei  and
      Wu, Hua  and
      Sun, Maosong  and
      Liu, Yang",
    booktitle = "Proceedings of the 54th Annual Meeting of the Association for Computational Linguistics (Volume 1: Long Papers)",
    month = aug,
    year = "2016",
    address = "Berlin, Germany",
    publisher = "Association for Computational Linguistics",
    url = "https://aclanthology.org/P16-1159",
    doi = "10.18653/v1/P16-1159",
    pages = "1683--1692",
}

@inproceedings{taivalkoski-shilov-2019-free,
    title = "Free indirect discourse: an insurmountable challenge for literary {MT} systems?",
    author = "Taivalkoski-Shilov, Kristiina",
    booktitle = "Proceedings of the Qualities of Literary Machine Translation",
    month = aug,
    year = "2019",
    address = "Dublin, Ireland",
    publisher = "European Association for Machine Translation",
    url = "https://aclanthology.org/W19-7305",
    pages = "35--39",
}

@inproceedings{voigt-jurafsky-2012-towards,
    title = "Towards a Literary Machine Translation: The Role of Referential Cohesion",
    author = "Voigt, Rob  and
      Jurafsky, Dan",
    booktitle = "Proceedings of the {NAACL}-{HLT} 2012 Workshop on Computational Linguistics for Literature",
    month = jun,
    year = "2012",
    address = "Montr{\'e}al, Canada",
    publisher = "Association for Computational Linguistics",
    url = "https://aclanthology.org/W12-2503",
    pages = "18--25",
}

@inproceedings{wieting-etal-2019-beyond,
    title = "Beyond {BLEU}:Training Neural Machine Translation with Semantic Similarity",
    author = "Wieting, John  and
      Berg-Kirkpatrick, Taylor  and
      Gimpel, Kevin  and
      Neubig, Graham",
    booktitle = "Proceedings of the 57th Annual Meeting of the Association for Computational Linguistics",
    month = jul,
    year = "2019",
    address = "Florence, Italy",
    publisher = "Association for Computational Linguistics",
    url = "https://aclanthology.org/P19-1427",
    doi = "10.18653/v1/P19-1427",
    pages = "4344--4355",
}

@inproceedings{wieting-etal-2019-simple,
    title = "Simple and Effective Paraphrastic Similarity from Parallel Translations",
    author = "Wieting, John  and
      Gimpel, Kevin  and
      Neubig, Graham  and
      Berg-Kirkpatrick, Taylor",
    booktitle = "Proceedings of the 57th Annual Meeting of the Association for Computational Linguistics",
    month = jul,
    year = "2019",
    address = "Florence, Italy",
    publisher = "Association for Computational Linguistics",
    url = "https://aclanthology.org/P19-1453",
    doi = "10.18653/v1/P19-1453",
    pages = "4602--4608",
}

@inproceedings{xue-etal-2021-mt5,
    title = "m{T}5: A Massively Multilingual Pre-trained Text-to-Text Transformer",
    author = "Xue, Linting  and
      Constant, Noah  and
      Roberts, Adam  and
      Kale, Mihir  and
      Al-Rfou, Rami  and
      Siddhant, Aditya  and
      Barua, Aditya  and
      Raffel, Colin",
    booktitle = "Proceedings of the 2021 Conference of the North American Chapter of the Association for Computational Linguistics: Human Language Technologies",
    month = jun,
    year = "2021",
    address = "Online",
    publisher = "Association for Computational Linguistics",
    url = "https://aclanthology.org/2021.naacl-main.41",
    doi = "10.18653/v1/2021.naacl-main.41",
    pages = "483--498",
}

@inproceedings{zheng-etal-2018-multi,
    title = "Multi-Reference Training with Pseudo-References for Neural Translation and Text Generation",
    author = "Zheng, Renjie  and
      Ma, Mingbo  and
      Huang, Liang",
    booktitle = "Proceedings of the 2018 Conference on Empirical Methods in Natural Language Processing",
    month = oct # "-" # nov,
    year = "2018",
    address = "Brussels, Belgium",
    publisher = "Association for Computational Linguistics",
    url = "https://aclanthology.org/D18-1357",
    doi = "10.18653/v1/D18-1357",
    pages = "3188--3197",
}

@inproceedings{wieting-etal-2022-paraphrastic,
    title = "Paraphrastic Representations at Scale",
    author = "Wieting, John  and
      Gimpel, Kevin  and
      Neubig, Graham  and
      Berg-kirkpatrick, Taylor",
    editor = "Che, Wanxiang  and
      Shutova, Ekaterina",
    booktitle = "Proceedings of the 2022 Conference on Empirical Methods in Natural Language Processing: System Demonstrations",
    month = dec,
    year = "2022",
    address = "Abu Dhabi, UAE",
    publisher = "Association for Computational Linguistics",
    url = "https://aclanthology.org/2022.emnlp-demos.38",
    doi = "10.18653/v1/2022.emnlp-demos.38",
    pages = "379--388",
}

@inproceedings{thai-etal-2022-exploring,
    title = "Exploring Document-Level Literary Machine Translation with Parallel Paragraphs from World Literature",
    author = "Thai, Katherine  and
      Karpinska, Marzena  and
      Krishna, Kalpesh  and
      Ray, Bill  and
      Inghilleri, Moira  and
      Wieting, John  and
      Iyyer, Mohit",
    editor = "Goldberg, Yoav  and
      Kozareva, Zornitsa  and
      Zhang, Yue",
    booktitle = "Proceedings of the 2022 Conference on Empirical Methods in Natural Language Processing",
    month = dec,
    year = "2022",
    address = "Abu Dhabi, United Arab Emirates",
    publisher = "Association for Computational Linguistics",
    url = "https://aclanthology.org/2022.emnlp-main.672",
    doi = "10.18653/v1/2022.emnlp-main.672",
    pages = "9882--9902",
}

@inproceedings{
hu2022lora,
title={Lo{RA}: Low-Rank Adaptation of Large Language Models},
author={Edward J Hu and yelong shen and Phillip Wallis and Zeyuan Allen-Zhu and Yuanzhi Li and Shean Wang and Lu Wang and Weizhu Chen},
booktitle={International Conference on Learning Representations},
year={2022},
url={https://openreview.net/forum?id=nZeVKeeFYf9}
}

@misc{su2022read,
      title={Read before Generate! Faithful Long Form Question Answering with Machine Reading}, 
      author={Dan Su and Xiaoguang Li and Jindi Zhang and Lifeng Shang and Xin Jiang and Qun Liu and Pascale Fung},
      year={2022},
      eprint={2203.00343},
      archivePrefix={arXiv},
      primaryClass={cs.CL}
}

@misc{horvitz2024paraguide,
      title={ParaGuide: Guided Diffusion Paraphrasers for Plug-and-Play Textual Style Transfer}, 
      author={Zachary Horvitz and Ajay Patel and Chris Callison-Burch and Zhou Yu and Kathleen McKeown},
      year={2024},
      eprint={2308.15459},
      archivePrefix={arXiv},
      primaryClass={cs.CL}
}

@book{nida1964toward,
  title={Toward a Science of Translating: With Special Reference to Principles and Procedures Involved in Bible Translating (Second edition)},
  author={Nida, E.A.},
  isbn={9789004495746},
  year={1964},
  publisher={Brill}
}

@book{andre,
  title={Translating literature : practice and theory in a comparative literature context},
  author={Lefevere, Andr\'e},
  isbn={978-0873523943},
  year={1992},
  publisher={New York :Modern Language Association of America}
}

@inproceedings{Jayawardena_2024, series={AIBD},
   title={Parafusion: A Large-Scale LLM-Driven English Paraphrase Dataset Infused with High-Quality Lexical and Syntactic Diversity},
   url={http://dx.doi.org/10.5121/csit.2024.140418},
   DOI={10.5121/csit.2024.140418},
   booktitle={Artificial Intelligence and Big Data},
   publisher={Academy \& Industry Research Collaboration Center},
   author={Jayawardena, Lasal and Yapa, Prasan},
   year={2024},
   month=feb, collection={AIBD} }

@inproceedings{Jayawardena_2024_gen,
   title={Parameter Efficient Diverse Paraphrase Generation Using Sequence-Level Knowledge Distillation},
   url={http://dx.doi.org/10.1109/ICACS60934.2024.10473289},
   DOI={10.1109/icacs60934.2024.10473289},
   booktitle={2024 5th International Conference on Advancements in Computational Sciences (ICACS)},
   publisher={IEEE},
   author={Jayawardena, Lasal and Yapa, Prasan},
   year={2024},
   month=feb }

@article{dettmers2023qlora,
  title={QLoRA: Efficient Finetuning of Quantized LLMs},
  author={Dettmers, Tim and Pagnoni, Artidoro and Holtzman, Ari and Zettlemoyer, Luke},
  journal={arXiv preprint arXiv:2305.14314},
  year={2023}
}

@misc{mishra2024finegrained,
      title={Fine-grained Hallucination Detection and Editing for Language Models}, 
      author={Abhika Mishra and Akari Asai and Vidhisha Balachandran and Yizhong Wang and Graham Neubig and Yulia Tsvetkov and Hannaneh Hajishirzi},
      year={2024},
      eprint={2401.06855},
      archivePrefix={arXiv},
      primaryClass={id='cs.CL' full_name='Computation and Language' is_active=True alt_name='cmp-lg' in_archive='cs' is_general=False description='Covers natural language processing. Roughly includes material in ACM Subject Class I.2.7. Note that work on artificial languages (programming languages, logics, formal systems) that does not explicitly address natural-language issues broadly construed (natural-language processing, computational linguistics, speech, text retrieval, etc.) is not appropriate for this area.'}
}

@inproceedings{rei-etal-2022-comet,
    title = "{COMET}-22: Unbabel-{IST} 2022 Submission for the Metrics Shared Task",
    author = "Rei, Ricardo  and
      C. de Souza, Jos{\'e} G.  and
      Alves, Duarte  and
      Zerva, Chrysoula  and
      Farinha, Ana C  and
      Glushkova, Taisiya  and
      Lavie, Alon  and
      Coheur, Luisa  and
      Martins, Andr{\'e} F. T.",
    editor = {Koehn, Philipp  and
      Barrault, Lo{\"\i}c  and
      Bojar, Ond{\v{r}}ej  and
      Bougares, Fethi  and
      Chatterjee, Rajen  and
      Costa-juss{\`a}, Marta R.  and
      Federmann, Christian  and
      Fishel, Mark  and
      Fraser, Alexander  and
      Freitag, Markus  and
      Graham, Yvette  and
      Grundkiewicz, Roman  and
      Guzman, Paco  and
      Haddow, Barry  and
      Huck, Matthias  and
      Jimeno Yepes, Antonio  and
      Kocmi, Tom  and
      Martins, Andr{\'e}  and
      Morishita, Makoto  and
      Monz, Christof  and
      Nagata, Masaaki  and
      Nakazawa, Toshiaki  and
      Negri, Matteo  and
      N{\'e}v{\'e}ol, Aur{\'e}lie  and
      Neves, Mariana  and
      Popel, Martin  and
      Turchi, Marco  and
      Zampieri, Marcos},
    booktitle = "Proceedings of the Seventh Conference on Machine Translation (WMT)",
    month = dec,
    year = "2022",
    address = "Abu Dhabi, United Arab Emirates (Hybrid)",
    publisher = "Association for Computational Linguistics",
    url = "https://aclanthology.org/2022.wmt-1.52",
    pages = "578--585",
}

@article{tiedemann2023democratizing,
  title={Democratizing neural machine translation with {OPUS-MT}},
  author={Tiedemann, J{\"o}rg and Aulamo, Mikko and Bakshandaeva, Daria and Boggia, Michele and Gr{\"o}nroos, Stig-Arne and Nieminen, Tommi and Raganato\
, Alessandro and Scherrer, Yves and Vazquez, Raul and Virpioja, Sami},
  journal={Language Resources and Evaluation},
  number={58},
  pages={713--755},
  year={2023},
  publisher={Springer Nature},
  issn={1574-0218},
  doi={10.1007/s10579-023-09704-w}
}

@InProceedings{TiedemannThottingal:EAMT2020,
  author = {J{\"o}rg Tiedemann and Santhosh Thottingal},
  title = {{OPUS-MT} — {B}uilding open translation services for the {W}orld},
  booktitle = {Proceedings of the 22nd Annual Conferenec of the European Association for Machine Translation (EAMT)},
  year = {2020},
  address = {Lisbon, Portugal}
 }

@Article{human-vs-NMT-paster,
AUTHOR = {Corpas Pastor, Gloria and Noriega-Santiáñez, Laura},
TITLE = {Human versus Neural Machine Translation Creativity: A Study on Manipulated MWEs in Literature},
JOURNAL = {Information},
VOLUME = {15},
YEAR = {2024},
NUMBER = {9},
ARTICLE-NUMBER = {530},
URL = {https://www.mdpi.com/2078-2489/15/9/530},
ISSN = {2078-2489},
DOI = {10.3390/info15090530}
}

@misc{castaldo2025extendingcreamtleveraginglarge,
      title={Extending CREAMT: Leveraging Large Language Models for Literary Translation Post-Editing}, 
      author={Antonio Castaldo and Sheila Castilho and Joss Moorkens and Johanna Monti},
      year={2025},
      eprint={2504.03045},
      archivePrefix={arXiv},
      primaryClass={cs.CL},
      url={https://arxiv.org/abs/2504.03045}, 
}

@incollection{brusasco2022pragmatic,
  title={Pragmatic and cognitive elements in literary machine translation: An assessment of an excerpt from J. Polzin's Brood translated with Google, DeepL, and Microsoft Bing},
  author={Brusasco, Paola},
  booktitle={Using Technologies for Creative-Text Translation},
  pages={139--160},
  year={2022},
  publisher={Routledge}
}

@inproceedings{li-etal-2024-linchance,
    title = "{L}in{C}hance-{NTU} for Unconstrained {WMT}2024 Literary Translation",
    author = "Li, Kechen  and
      Tao, Yaotian  and
      Huang, Hongyi  and
      Ji, Tianbo",
    editor = "Haddow, Barry  and
      Kocmi, Tom  and
      Koehn, Philipp  and
      Monz, Christof",
    booktitle = "Proceedings of the Ninth Conference on Machine Translation",
    month = nov,
    year = "2024",
    address = "Miami, Florida, USA",
    publisher = "Association for Computational Linguistics",
    url = "https://aclanthology.org/2024.wmt-1.99/",
    doi = "10.18653/v1/2024.wmt-1.99",
    pages = "987--992"
}

@misc{gemmateam2025gemma3technicalreport,
      title={Gemma 3 Technical Report}, 
      author={Gemma Team and Aishwarya Kamath and Johan Ferret and Shreya Pathak and Nino Vieillard and Ramona Merhej and Sarah Perrin and Tatiana Matejovicova and Alexandre Ramé and Morgane Rivière and Louis Rouillard and Thomas Mesnard and Geoffrey Cideron and Jean-bastien Grill and Sabela Ramos and Edouard Yvinec and Michelle Casbon and Etienne Pot and Ivo Penchev and Gaël Liu and Francesco Visin and Kathleen Kenealy and Lucas Beyer and Xiaohai Zhai and Anton Tsitsulin and Robert Busa-Fekete and Alex Feng and Noveen Sachdeva and Benjamin Coleman and Yi Gao and Basil Mustafa and Iain Barr and Emilio Parisotto and David Tian and Matan Eyal and Colin Cherry and Jan-Thorsten Peter and Danila Sinopalnikov and Surya Bhupatiraju and Rishabh Agarwal and Mehran Kazemi and Dan Malkin and Ravin Kumar and David Vilar and Idan Brusilovsky and Jiaming Luo and Andreas Steiner and Abe Friesen and Abhanshu Sharma and Abheesht Sharma and Adi Mayrav Gilady and Adrian Goedeckemeyer and Alaa Saade and Alex Feng and Alexander Kolesnikov and Alexei Bendebury and Alvin Abdagic and Amit Vadi and András György and André Susano Pinto and Anil Das and Ankur Bapna and Antoine Miech and Antoine Yang and Antonia Paterson and Ashish Shenoy and Ayan Chakrabarti and Bilal Piot and Bo Wu and Bobak Shahriari and Bryce Petrini and Charlie Chen and Charline Le Lan and Christopher A. Choquette-Choo and CJ Carey and Cormac Brick and Daniel Deutsch and Danielle Eisenbud and Dee Cattle and Derek Cheng and Dimitris Paparas and Divyashree Shivakumar Sreepathihalli and Doug Reid and Dustin Tran and Dustin Zelle and Eric Noland and Erwin Huizenga and Eugene Kharitonov and Frederick Liu and Gagik Amirkhanyan and Glenn Cameron and Hadi Hashemi and Hanna Klimczak-Plucińska and Harman Singh and Harsh Mehta and Harshal Tushar Lehri and Hussein Hazimeh and Ian Ballantyne and Idan Szpektor and Ivan Nardini and Jean Pouget-Abadie and Jetha Chan and Joe Stanton and John Wieting and Jonathan Lai and Jordi Orbay and Joseph Fernandez and Josh Newlan and Ju-yeong Ji and Jyotinder Singh and Kat Black and Kathy Yu and Kevin Hui and Kiran Vodrahalli and Klaus Greff and Linhai Qiu and Marcella Valentine and Marina Coelho and Marvin Ritter and Matt Hoffman and Matthew Watson and Mayank Chaturvedi and Michael Moynihan and Min Ma and Nabila Babar and Natasha Noy and Nathan Byrd and Nick Roy and Nikola Momchev and Nilay Chauhan and Noveen Sachdeva and Oskar Bunyan and Pankil Botarda and Paul Caron and Paul Kishan Rubenstein and Phil Culliton and Philipp Schmid and Pier Giuseppe Sessa and Pingmei Xu and Piotr Stanczyk and Pouya Tafti and Rakesh Shivanna and Renjie Wu and Renke Pan and Reza Rokni and Rob Willoughby and Rohith Vallu and Ryan Mullins and Sammy Jerome and Sara Smoot and Sertan Girgin and Shariq Iqbal and Shashir Reddy and Shruti Sheth and Siim Põder and Sijal Bhatnagar and Sindhu Raghuram Panyam and Sivan Eiger and Susan Zhang and Tianqi Liu and Trevor Yacovone and Tyler Liechty and Uday Kalra and Utku Evci and Vedant Misra and Vincent Roseberry and Vlad Feinberg and Vlad Kolesnikov and Woohyun Han and Woosuk Kwon and Xi Chen and Yinlam Chow and Yuvein Zhu and Zichuan Wei and Zoltan Egyed and Victor Cotruta and Minh Giang and Phoebe Kirk and Anand Rao and Kat Black and Nabila Babar and Jessica Lo and Erica Moreira and Luiz Gustavo Martins and Omar Sanseviero and Lucas Gonzalez and Zach Gleicher and Tris Warkentin and Vahab Mirrokni and Evan Senter and Eli Collins and Joelle Barral and Zoubin Ghahramani and Raia Hadsell and Yossi Matias and D. Sculley and Slav Petrov and Noah Fiedel and Noam Shazeer and Oriol Vinyals and Jeff Dean and Demis Hassabis and Koray Kavukcuoglu and Clement Farabet and Elena Buchatskaya and Jean-Baptiste Alayrac and Rohan Anil and Dmitry and Lepikhin and Sebastian Borgeaud and Olivier Bachem and Armand Joulin and Alek Andreev and Cassidy Hardin and Robert Dadashi and Léonard Hussenot},
      year={2025},
      eprint={2503.19786},
      archivePrefix={arXiv},
      primaryClass={cs.CL},
      url={https://arxiv.org/abs/2503.19786}, 
}

@incollection{jakobson1987linguistics,
  author    = {Jakobson, Roman},
  title     = {Linguistics and Poetics},
  booktitle = {Language in Literature},
  editor    = {Pomorska, Krystyna and Rudy, Stephen},
  pages     = {62--94},
  publisher = {Harvard University Press},
  address   = {Cambridge, MA},
  year      = {1987}
}

@InProceedings{smith06:minrisk,
  author =	 {David A. Smith and Jason Eisner},
  title =	 {Minimum Risk Annealing for Training Log-Linear
                  Models},
  sortmonth = 	 07,
  booktitle =	 COLING-ACL,
  year =	 2006,
  pages = 	 {787-794}
}

@article{nabokov1955problems,
  author  = {Nabokov, Vladimir},
  title   = {Problems of Translation: {``Onegin''} in English},
  journal = {Partisan Review},
  volume  = {22},
  number  = {4},
  year    = {1955},
  pages   = {496--512}
}

@book{hofstadter1997music,
  author    = {Hofstadter, Douglas R.},
  title     = {Le Ton beau de Marot: In Praise of the Music of Language},
  publisher = {Basic Books},
  address   = {New York},
  year      = {1997}
}

@article{szymanska2026manifesto,
title = "A manifesto on multiple translation: Bridging retranslation, world literature, and the ethics of comparative reading",
author  = {Szymanska, Kasia},
year = "2026",
month = apr,
day = "20",
doi = "10.1080/14781700.2026.2625904",
language = "English",
journal = "Translation Studies",
issn = "1478-1700",
publisher = "Taylor \& Francis",
}

@book{rose1997translation,
  author    = {Rose, Marilyn Gaddis},
  title     = {Translation and Literary Criticism: Translation as Analysis},
  publisher = {St. Jerome Publishing},
  year      = {1997},
  address   = {Manchester, UK}
}

@article{stenberg2018multiplicitypoetry,
  author  = {Stenberg, Josh},
  title   = {Multiplicity in Lieu of Authority: Translations of Classical Chinese Poetry Online},
  journal = {Babel},
  year    = {2018},
  volume  = {64},
  number  = {4},
  pages   = {579--593},
  doi     = {10.1075/babel.00058.ste}
}

@inproceedings{de-gibert-etal-2025-scaling,
    title = "Scaling Low-Resource {MT} via Synthetic Data Generation with {LLM}s",
    author = {de Gibert, Ona  and
      Attieh, Joseph  and
      Vahtola, Teemu  and
      Aulamo, Mikko  and
      Li, Zihao  and
      V{\'a}zquez, Ra{\'u}l  and
      Hu, Tiancheng  and
      Tiedemann, J{\"o}rg},
    editor = "Christodoulopoulos, Christos  and
      Chakraborty, Tanmoy  and
      Rose, Carolyn  and
      Peng, Violet",
    booktitle = "Proceedings of the 2025 Conference on Empirical Methods in Natural Language Processing",
    month = nov,
    year = "2025",
    address = "Suzhou, China",
    publisher = "Association for Computational Linguistics",
    url = "https://aclanthology.org/2025.emnlp-main.1408/",
    doi = "10.18653/v1/2025.emnlp-main.1408",
    pages = "27674--27692",
    ISBN = "979-8-89176-332-6"
}

@inproceedings{poncelas-etal-2018-investigating,
    title = "Investigating Backtranslation in Neural Machine Translation",
    author = "Poncelas, Alberto  and
      Shterionov, Dimitar  and
      Way, Andy  and
      Maillette de Buy Wenniger, Gideon  and
      Passban, Peyman",
    editor = "P{\'e}rez-Ortiz, Juan Antonio  and
      S{\'a}nchez-Mart{\'i}nez, Felipe  and
      Espl{\`a}-Gomis, Miquel  and
      Popovi{\'c}, Maja  and
      Rico, Celia  and
      Martins, Andr{\'e}  and
      Van den Bogaert, Joachim  and
      Forcada, Mikel L.",
    booktitle = "Proceedings of the 21st Annual Conference of the European Association for Machine Translation",
    month = may,
    year = "2018",
    address = "Alicante, Spain",
    url = "https://aclanthology.org/2018.eamt-main.25/",
    pages = "269--278",
}

@inproceedings{long-etal-2024-llms,
    title = "On {LLM}s-Driven Synthetic Data Generation, Curation, and Evaluation: A Survey",
    author = "Long, Lin  and
      Wang, Rui  and
      Xiao, Ruixuan  and
      Zhao, Junbo  and
      Ding, Xiao  and
      Chen, Gang  and
      Wang, Haobo",
    editor = "Ku, Lun-Wei  and
      Martins, Andre  and
      Srikumar, Vivek",
    booktitle = "Findings of the Association for Computational Linguistics: ACL 2024",
    month = aug,
    year = "2024",
    address = "Bangkok, Thailand",
    publisher = "Association for Computational Linguistics",
    url = "https://aclanthology.org/2024.findings-acl.658/",
    doi = "10.18653/v1/2024.findings-acl.658",
    pages = "11065--11082"
}

@inproceedings{scalvini-etal-2025-rethinking,
    title = "Rethinking Low-Resource {MT:} The Surprising Effectiveness of Fine-Tuned Multilingual Models in the {LLM} Age",
    author = "Scalvini, Barbara  and
      Debess, Iben Nyholm  and
      Simonsen, Annika  and
      Einarsson, Hafsteinn",
    editor = "Johansson, Richard  and
      Stymne, Sara",
    booktitle = "Proceedings of the Joint 25th Nordic Conference on Computational Linguistics and 11th Baltic Conference on Human Language Technologies (NoDaLiDa/Baltic-HLT 2025)",
    month = mar,
    year = "2025",
    address = "Tallinn, Estonia",
    publisher = "University of Tartu Library",
    url = "https://aclanthology.org/2025.nodalida-1.62/",
    pages = "609--621",
    ISBN = "978-9908-53-109-0"
}

@inproceedings{mikelenic-etal-2025-fine,
    title = "Fine-tuning and evaluation of {NMT} models for literary texts using {R}om{C}ro v.2.0",
    author = "Mikeleni{\'c}, Bojana  and
      Oliver, Antoni  and
      Vidal, Sergi {\`A}lvarez",
    editor = "Vanroy, Bram  and
      Lefer, Marie-Aude  and
      Macken, Lieve  and
      Ruffo, Paola  and
      Arenas, Ana Guerberof  and
      Hansen, Damien",
    booktitle = "Proceedings of the Second Workshop on Creative-text Translation and Technology (CTT)",
    month = jun,
    year = "2025",
    address = "Geneva, Switzerland",
    publisher = "European Association for Machine Translation",
    url = "https://aclanthology.org/2025.ctt-1.4/",
    pages = "44--51",
    ISBN = "978-2-9701897-6-3"
}

@inproceedings{steingrimsson-etal-2023-filtering,
    title = "Filtering Matters: Experiments in Filtering Training Sets for Machine Translation",
    author = "Steingr{\'i}msson, Stein{\th}{\'o}r  and
      Loftsson, Hrafn  and
      Way, Andy",
    editor = {Alum{\"a}e, Tanel  and
      Fishel, Mark},
    booktitle = "Proceedings of the 24th Nordic Conference on Computational Linguistics (NoDaLiDa)",
    month = may,
    year = "2023",
    address = "T{\'o}rshavn, Faroe Islands",
    publisher = "University of Tartu Library",
    url = "https://aclanthology.org/2023.nodalida-1.58/",
    pages = "588--600"
}

@inproceedings{kwon2023efficient,
  title={Efficient Memory Management for Large Language Model Serving with PagedAttention},
  author={Woosuk Kwon and Zhuohan Li and Siyuan Zhuang and Ying Sheng and Lianmin Zheng and Cody Hao Yu and Joseph E. Gonzalez and Hao Zhang and Ion Stoica},
  booktitle={Proceedings of the ACM SIGOPS 29th Symposium on Operating Systems Principles},
  year={2023}
}
\bibliographystyle{acl_natbib}

\newpage
\appendix


\section{Pre-Processing} \label{pre-processing}

\begin{figure*}[t]
\centering
\includegraphics[width=0.45\linewidth]{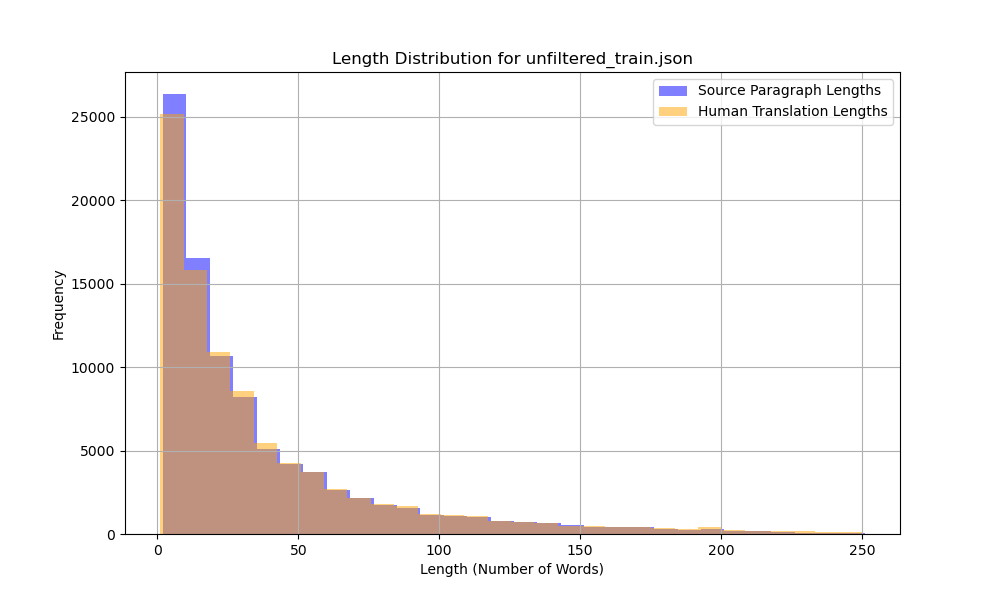}
\hspace{0.05\linewidth}
\includegraphics[width=0.45\linewidth]{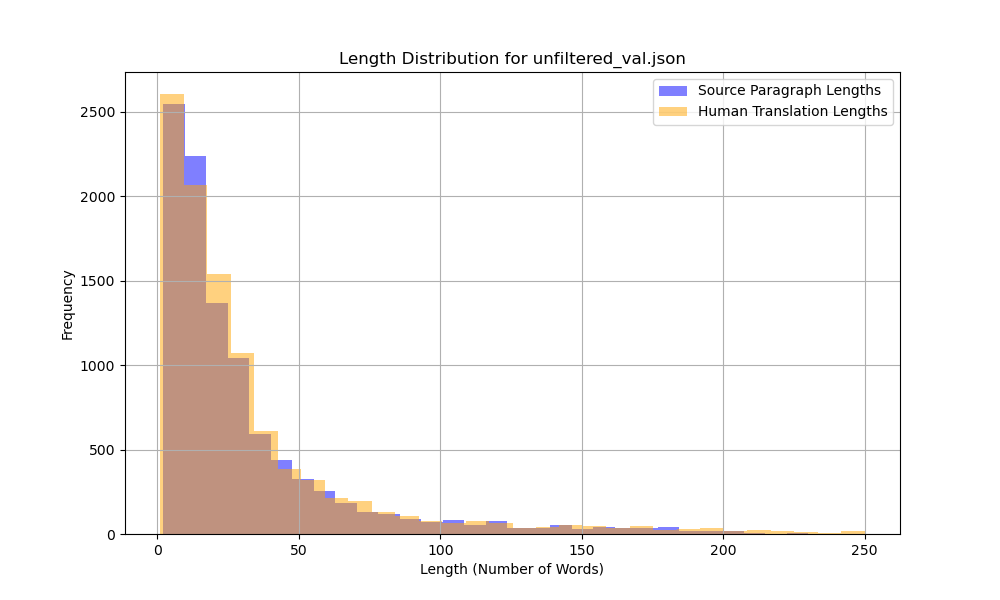}
\caption{Distribution of paragraph word lengths in the French TRAIN (left) and VAL (right) datasets. Each instance represents a source paragraph or translation instance, showing the frequency of paragraphs by word count.}
\label{fig-para-length}
\end{figure*}

We filter out source paragraphs and translations longer than 250 words to remove extreme outliers and ensure reliable processing by sentence-based MT models like OpusMT. The 250-word cutoff is also informed by the distribution of word length in the dataset: in Figure~\ref{fig-para-length}, we show the distributions of both source and target paragraphs length measured by word count in the French train and validation datasets.  After filtering, we only keep instances where each source paragraph has two or more human translations.

Then, for each language, we construct train, validation, and test splits at the book level, preventing any paragraph overlap across splits. 
We target an 80/10/10 paragraph ratio. Exact statistics for training instances for all splits and datasets are reported in Table~\ref{low_med_stats_fr} and \ref{low_med_stats_ru}.

The train and validation datasets are the starting points for more targeted experiments from Sections~\ref{Section-finding-optimal} to \ref{Section-synthetic}, where we will further filter by the semantic similarity among the human translations. The test dataset created is used to test all experiments for a specific language pair. The French test set contains 5,030 source texts, and Russian has 3,292. Each source text in the test set has multiple human expert translations as targets.

\begin{table}[h]
\centering
\resizebox{\linewidth}{!}{%

\centering
\small
\begin{tabular}{lcccc}
\toprule
& \multicolumn{2}{c}{\textbf{Train}} 
& \multicolumn{2}{c}{\textbf{Validation}} \\
\cmidrule(lr){2-3} \cmidrule(lr){4-5}
\textbf{Setting} 
& \textbf{src} & \textbf{instance}
& \textbf{src} & \textbf{instance} \\
\midrule
low         & 5,016 & 11,279  & 470  & 940 \\
med         & 4,977 & 11,279  & 470 & 940 \\
high        & 4,779 & 11,279  & 470 & 940 \\
\midrule
med+low     & 30,580 & 69,575 & 4,208 & 8,416 \\
med+high    &  30,595 & 69,575 & 4,208 & 8,416 \\
unfiltered    &  30,580 & 69,575 & 4,208 &  8,416\\
\midrule
unfiltered  & 39,880 & 91,647 & 4,987 & 9,974 \\
medium+high all  & 34,864 & 80,368 & 4,517  & 9,034 \\
\midrule
human (single ref)  & 34,864 & 34,864 & 4,517 & 4,517 \\
synthetic (Gemma3)  & 34,864 & 80,368 & 4,517 &  9,034\\
synthetic (Mistral)  & 34,864 & 80,368 & 4,517 & 9,034  \\
synthetic (Qwen3)  & 34,864 & 80,368 & 4,517 & 9,034 \\
\bottomrule
\end{tabular}
\label{tab:data_stats}

}
\caption{French train and validation source and instance counts.}
\label{low_med_stats_fr}
\end{table}

\begin{table}[h]
\centering
\resizebox{\linewidth}{!}{%
\centering
\small
\begin{tabular}{lcccc}
\toprule
& \multicolumn{2}{c}{\textbf{Train}} 
& \multicolumn{2}{c}{\textbf{Validation}} \\
\cmidrule(lr){2-3} \cmidrule(lr){4-5}
\textbf{Setting} 
& \textbf{src} & \textbf{instance}
& \textbf{src} & \textbf{instance} \\
\midrule
low         & 1,618 & 4,114  & 384 & 1,087 \\
med         & 1,529 & 4,114 & 506 &  1,087\\
high        & 1,874 & 4,114  & 520 & 1,087 \\
\midrule
med+low     & 17,530 & 47,641 & 2,322 & 5,491 \\
med+high    & 17,969 & 47,641 & 2,534 & 2,534 \\
unfiltered    & 18,744 & 47,641 & 2,344 & 5,491  \\
\midrule
unfiltered  & 26,502 & 67,106  & 3,296 & 7,675 \\
medium+high all  & 24,884 &  62,992 & 2,534 & 5,491 \\
\midrule
human (single ref)  & 24,884  & 24,884 &  2,534 &  2,534 \\
synthetic (Gemma3)  & 24,884 & 62,992  & 2,534 & 5,491 \\
synthetic (Mistral)  & 24,884 &  62,992  & 2,534 & 5,491 \\
synthetic (Qwen3)  &24,884  &  62,992 & 2,534 & 5,491 \\
\bottomrule
\end{tabular}
\label{tab:data_stats}
}
\caption{Russian train and validation source and instance counts.}
\label{low_med_stats_ru}
\end{table}

\section{MT Models and Automatic Metrics} \label{appendix-model-metrics}

\subsection{Models}
\paragraph{mT5-large} A multilingual encoder--decoder Transformer model with approximately 1.2B parameters. It is pre-trained on the mC4 corpus, which covers 101 languages, including French and Russian \citep{xue-etal-2021-mt5}.\footnote{This model is available on Hugging Face at \url{https://huggingface.co/google/mt5-large}} Although mT5-large is not specifically pre-trained for translation, it can be fine-tuned for MT and other downstream tasks. We fine-tuned this model using the Hugging Face \texttt{Trainer} class. Each experiment takes less than 24 hours on a single NVIDIA H200 GPU. The learning rate $= 1e-3$,  batch size $= 4$, and gradient accumulation $= 4$.

\paragraph{Gemma-3-12B-PT}
An open-weight, non-instruction-tuned language model with 12 billion parameters.\footnote{\url{https://huggingface.co/google/gemma-3-12b-pt}} Given its scale and our computational constraints, we adapt the model using LoRA with 4-bit quantization \cite{hu2022lora, dettmers2023qlora}. Gemma3 is trained on multilingual web text covering 140 languages, although the language distribution is not publicly specified \citep{gemmateam2025gemma3technicalreport}. We use LoRA rank $r=32$ and scaling parameter $\alpha=32$. Each experiment takes from 48 to 120 hours on a single NVIDIA H200 GPU, depending on the size of the dataset. The learning rate$= 1e-4$, batch size $= 4$, and gradient accumulation $= 8$.

\paragraph{OpusMT} An open-source multilingual machine translation model from the OPUS-MT project \citep{tiedemann2023democratizing, TiedemannThottingal:EAMT2020}.\footnote{\url{https://huggingface.co/Helsinki-NLP/opus-mt-mul-en}} The model supports multiple source languages with English as the target language and is trained on the OPUS corpus using the Marian framework \citep{junczys-dowmunt-etal-2018-marian}. Each experiment takes less than 12 hours on a single NVIDIA Tesla V100 SXM2 GPU. The learning rate $= 5e-5$, batch size $= 8$, and gradient accumulation $= 4$.

\subsection{Automatic Evaluation Metrics} \label{section-auto-metrics}

\paragraph{BLEU} An MT evaluation metric based on modified n-gram precision with a brevity penalty \citep{papineni-etal-2002-bleu}. We compute BLEU using multiple human references for each source input in the test set. 

\paragraph{$\text{COMET}_{22}^{\text{DA}}$} An MT evaluation metric that is based on sentence embeddings \citep{rei-etal-2020-comet}. We use wmt22-comet-da by Unbabel \citep{rei-etal-2022-comet}.\footnote{\url{https://huggingface.co/Unbabel/wmt22-comet-da}} It was trained on a transformer-based multilingual masked language model, XLM-R \citep{conneau-etal-2020-unsupervised}. Since $\text{COMET}_{22}^{\text{DA}}$ only accepts one reference per source text for evaluation, for each source text, we take the average of their $\text{COMET}_{22}^{\text{DA}}$ scores for each reference.

\paragraph{chrF++} An MT evaluation metric based on character and word-level n-gram precision and recall \citep{popovic-2017-chrf}.


 \newpage
\section{Models and Prompts Used for Generating Synthetic Data} \label{appendix-synthetic-model-prompts}

We first randomly select one human translation from each source text in the  full medium+high dataset. Then, each one of these human translations is used to generate paraphrase with the following models. The following models were all able to zero-shot generate paraphrases for at least 99\% of the data. 

We use 1,000 as the limit for max new token, and retry with 1,200 if the output is cut off. For the few instances where we could not successfully parse the model output because the model did not follow the instructions, we salvaged as many generated paraphrases as possible, then replaced any missing paraphrases with human translations.

\paragraph{gemma-3-27b-it\footnote{Available to download on Hugging Face at \url{https://huggingface.co/google/gemma-3-27b-it}}:}
gemma-3-27b-it is Google's 27-billion-parameter instruction-tuned Gemma3 model (different from the 12b-pt variant we use for fine-tuning). It is designed for instruction following, and it can handle long-context window. To generate text more efficiently, we perform zero-shot text generation using vLLM \citep{kwon2023efficient} with a batch size of 16. For the French TRAIN split, generation completed on a single H200 GPU in under 17 hours. We obtained outputs for all 34,864 instances. Only 2 instances produced a single paraphrase rather than the requested two. For the validation split, we successfully processed all 4,517 instances, with no single-paraphrase outputs and no discarded instances.

\begin{quote}
\small
System prompt: 
You are a helpful assistant.

User prompt:
Generate two paraphrases of the paragraph below. 
Each paraphrase should preserve the same meaning as the original, 
but express it in a different way. 
The two paraphrases should not be identical or trivial rewrites. 
Set \texttt{reference} to exactly the paragraph provided. 
Return ONLY valid JSON in the following format:

\begin{verbatim}
{
  "reference": "...",
  "paraphrase_1": "...",
  "paraphrase_2": "..."
}
\end{verbatim}

\texttt{[INPUT]}

\end{quote}

\paragraph{Qwen3-30B-A3B-Instruct-2507\footnote{Available to download on Hugging Face at \url{https://huggingface.co/Qwen/Qwen3-30B-A3B-Instruct-2507}}:}
An instruction-tuned mixture-of-experts language model from the Qwen3 family, with 30.5 billion total parameters and 3.3 billion activated parameters. It has strong instruction following and general reasoning capabilities. Like Gemma3, we ran inference using vLLM with a batch size of 8. The TRAIN split finished under 5 hours. Of the 34,864 source paragraphs in the TRAIN split, only 54 outputs contained a single paraphrase rather than the requested two paraphrases. For the validation split, all outputs were successfully salvaged; only 1 instance contained a single paraphrase.

\begin{quote}
\small
User prompt:
Generate two paraphrases of the paragraph below. 
Each paraphrase should convey the same meaning as the original, 
but be expressed in a different way. 
The two paraphrases should not be identical or trivial rewrites. 
Set \texttt{reference} to be exactly the paragraph provided, 
without any changes. 
Return only valid JSON in the following format:

\begin{verbatim}
{
  "reference": "...",
  "paraphrase_1": "...",
  "paraphrase_2": "..."
}
\end{verbatim}

\texttt{[INPUT]}

\end{quote}

\paragraph{Mistral-Small-3.2-24B-Instruct-2506\footnote{Available to download on Hugging Face at \url{https://huggingface.co/mistralai/Mistral-Small-3.2-24B-Instruct-2506}}:}
 A 24-billion-parameter instruction-tuned model from Mistral AI. It is designed for efficient instruction following as well as long context use. We ran inference using vLLM with a batch size of 8. The TRAIN split finished in under 24 hours. For the TRAIN split, 85 outputs could not be salvaged. 1,959 instances had one single paraphrase. For the validation split, all 4,517 instances were processed: 4,290 produced complete paraphrase pairs, 220 produced a single recoverable paraphrase, and 7 were discarded and replaced with human translations.

\begin{quote}
\small
User prompt:
Generate two paraphrases of the paragraph below. 
Each paraphrase must preserve the original meaning but be written differently. 
Avoid trivial rewrites.

Return ONLY valid JSON in the following format:

\begin{verbatim}
{
  "reference": "...",
  "paraphrase_1": "...",
  "paraphrase_2": "..."
}
\end{verbatim}

\texttt{[INPUT]}

\end{quote}


 \newpage
\section{Zero-shot Translation with Instruction-tuned Models} \label{section-zero-shot}

While this paper primarily explores fine-tuning MT models using datasets with different levels of semantic similarity using multi-reference data, it is also informative to know the zero-shot translation performance of commercial models (via API) and open-weight instruction-tuned LLMs. In Table~\ref{tab:zero-shot-results} we show their performance measured by automatic metrics on the French test set. 

Results from the automatic metrics show that while commercial models like GPT 5.4 Pro outperform our best model (medium+high (all) with mT5) by 4.2 in BLEU, 0.01 in COMET, and 0.04 in chrF++, other state-of-the-art commercial models like Gemini 3.1 pro and Claude Opus 4.6 are only marginally better or worse than our best model. 

Open-weight models not only all underperform our best fine-tuned model, some even struggle to process and translate given the instructions, with a valid output rate as low as 52.9\% using Mistral-Small-3.2-24B-Instruct-2506 (Table~\ref{tab:zero-shot-results}).

\begin{table}[h]
\centering
\resizebox{\linewidth}{!}{%
\begin{tabular}{l ccccc}
\toprule
\textbf{model} & \textbf{cost} & \textbf{BLEU} & \textbf{COMET} & \textbf{chrF++} & \textbf{Valid\%}\\
\midrule
GPT 5.4 pro & \$15.47 & \textbf{48.5} & \textbf{0.76} & \textbf{0.6} & 100 \\
Gemini 3.1 Pro & \$37 & \textbf{44.8} &  \textbf{0.76} & 0.55 & 99.9\\
Claude Opus 4.6  & \$7.95 & 44.2 & \textbf{0.76} & \textbf{0.6} & 100\\
\midrule
Gemma3-27b-instruct  & 0  & 40.6   &   0.75    &  0.55  & 99.5 \\
Qwen3-instruct & 0  & 40.1   &  0.73 & 0.54    & 75.9  \\
Mistral-instruct  & 0 & 35.4    &  0.66    & 0.50 & 52.9 \\
\midrule
\midrule
\textbf{our best FT} & 0 & 44.3    &  0.75    & 0.56 & 100 \\
\bottomrule
\end{tabular}

}
\caption{Zero-shot machine translation performance on French-English test data. Scores in bold font are better than our best fine-tuned model---fine-tuning mT5 using the full medium and high \semsim data. ``Valid\%'' counts how many test data were able to be processed and translated. Cost for Gemini is estimated. }
\label{tab:zero-shot-results}
\end{table}

Below, we provide details of these models.

\subsection{Commercial Models}

\paragraph{GPT 5.4 Pro} A commercial model developed by OpenAI with long context window, multilingual capability, and instruction following ability. We used the model with reasoning effort set to \texttt{high}.

\paragraph{Gemini 3.1 pro} A commercial model developed by Google Deepmind. Similar to GPT 5.4 Pro, it has multilingual generation capability, long context window, and instruction following ability. We used the model with reasoning effort set to \texttt{high}.

\paragraph{Claude Opus 4.6} A commercial model developed by Anthropic. It is capable of multilingual tasks, has a massive 1 million token context window, and instruction following ability.

For all commercial models, we use the following system prompt for zero-shot machine translation. All APIs were called in batch mode. 

\begin{quote}
\small
System prompt: 
You are an expert literary translator from French to English. 
Translate the following French paragraph into natural, fluent English 
that preserves the original tone, style, and nuance. 
Output ONLY the English translation with no explanation or commentary.

\texttt{[INPUT]}

\end{quote}

\subsection{Open-Weight Models}

For zero-shot translation, we used the same open-weight models that we used to zero-shot generate paraphrase in Appendix~\ref{appendix-synthetic-model-prompts}. Below, we show the prompts we used for generation.

\paragraph{gemma-3-27b-it} 
\begin{quote}
\small
System prompt:
You are a careful literary translator.

User prompt:\\
Task: Translate the following French paragraph into natural, fluent English. \\
Style: literary; preserve tone, register, and meaning; avoid adding new facts. \\
Output format: Return ONLY a valid JSON object with exactly this key:

\begin{verbatim}
{
  "prediction": "..."
}
\end{verbatim}

Rules:
Do NOT include markdown, code fences, commentary, or extra keys. \\
Keep the English as a single string, including punctuation or quotes as needed.

French paragraph:

\texttt{[INPUT]}

\end{quote}

\paragraph{Qwen3-30B-A3B-Instruct-2507} 
\begin{quote}
\small
User prompt: \\
Task: Translate the following French paragraph into natural, fluent English. \\
Style: literary; preserve tone, register, and meaning; do not add new facts. \\

Return ONLY valid JSON with exactly this key: \\
\begin{verbatim}
{
  "prediction": "..."
}
\end{verbatim}
Rules: \\
- No markdown, no code fences, no explanations. \\
- Do NOT add extra keys. \\
- Keep the English as a single string. \\[0.5em]

French paragraph: \\
\texttt{[INPUT]}

\end{quote}

\paragraph{Mistral-Small-3.2-24B-Instruct-2506} 
\begin{quote}
\small
User prompt: \\
Task: Translate the following French paragraph into natural, fluent English. \\
Style: literary; preserve tone, register, and meaning; do not add new facts. \\[0.5em]

Return ONLY valid JSON with exactly this key: \\
\begin{verbatim}
{
  "prediction": "..."
}
\end{verbatim}
Rules: \\
- No markdown, no code fences, no extra keys, no explanations. \\
- Keep the English as a single string. \\[0.5em]

French paragraph: \\
\texttt{[INPUT]}

\end{quote}

\newpage
\section{\spavgfull Score Distribution of the Synthetic Data} \label{appendix-psp}
\begin{figure}[h]
\centering
\includegraphics[width=0.49\linewidth]{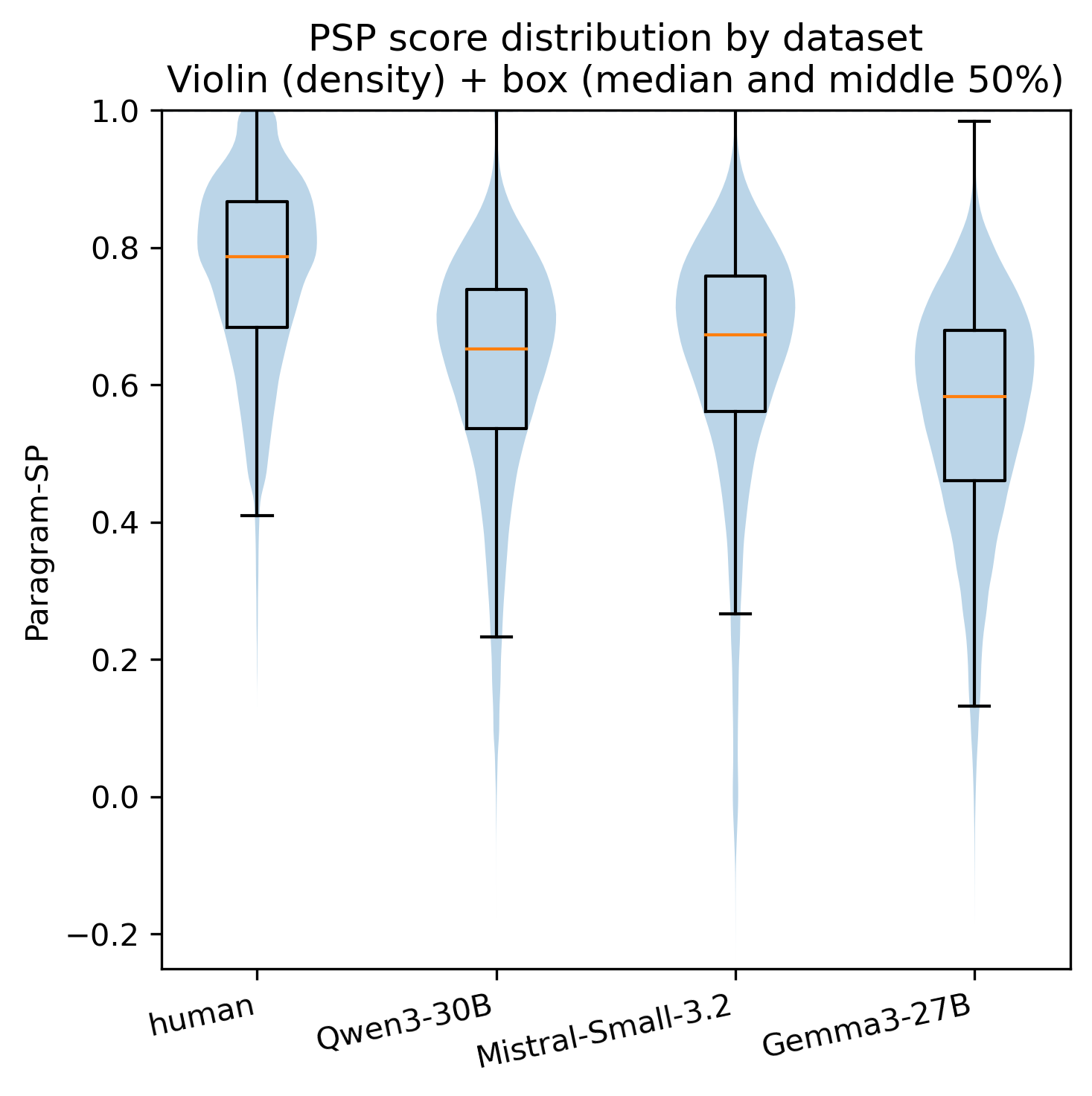}
\includegraphics[width=0.49\linewidth]{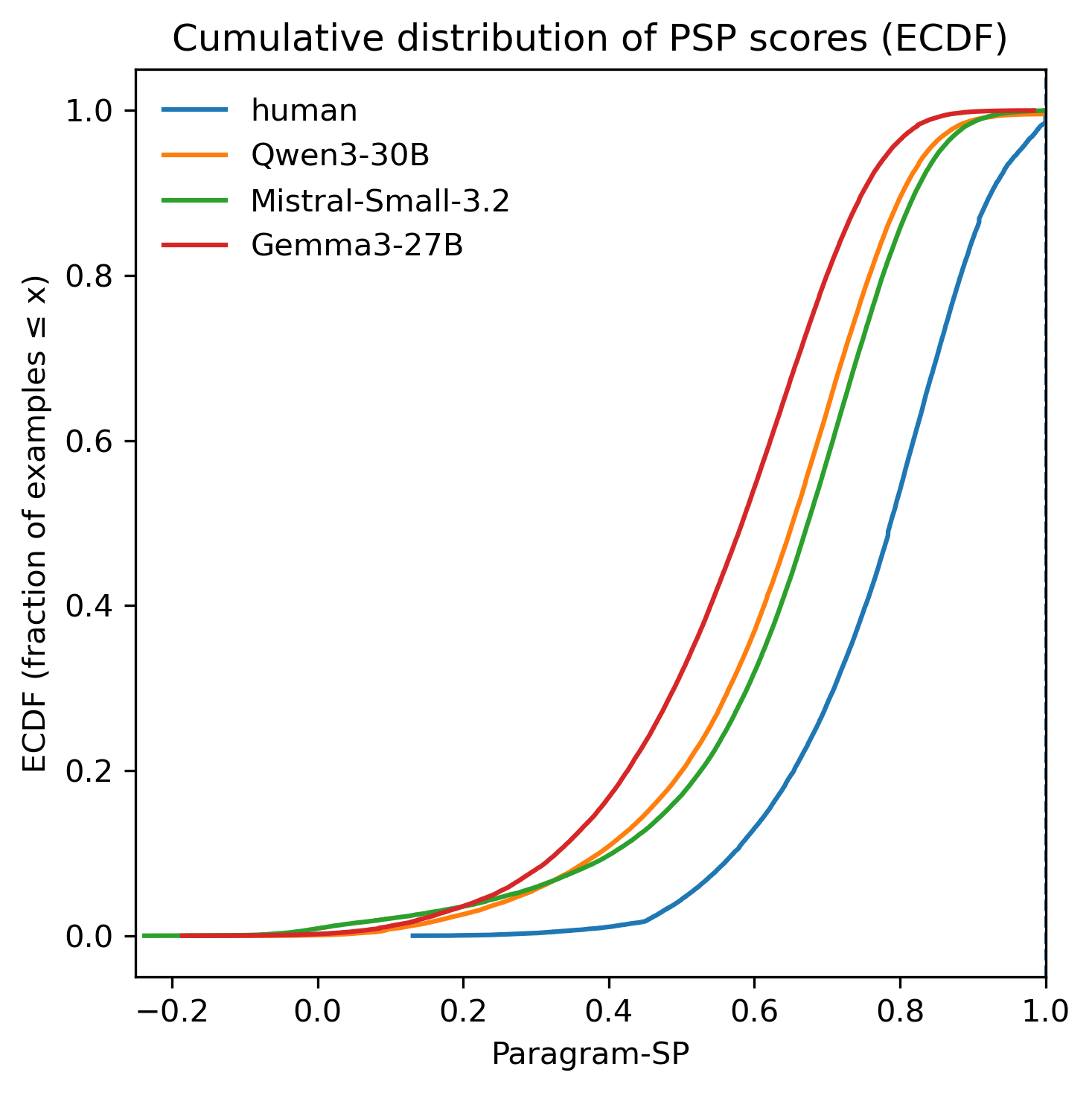}
\includegraphics[width=0.49\linewidth]{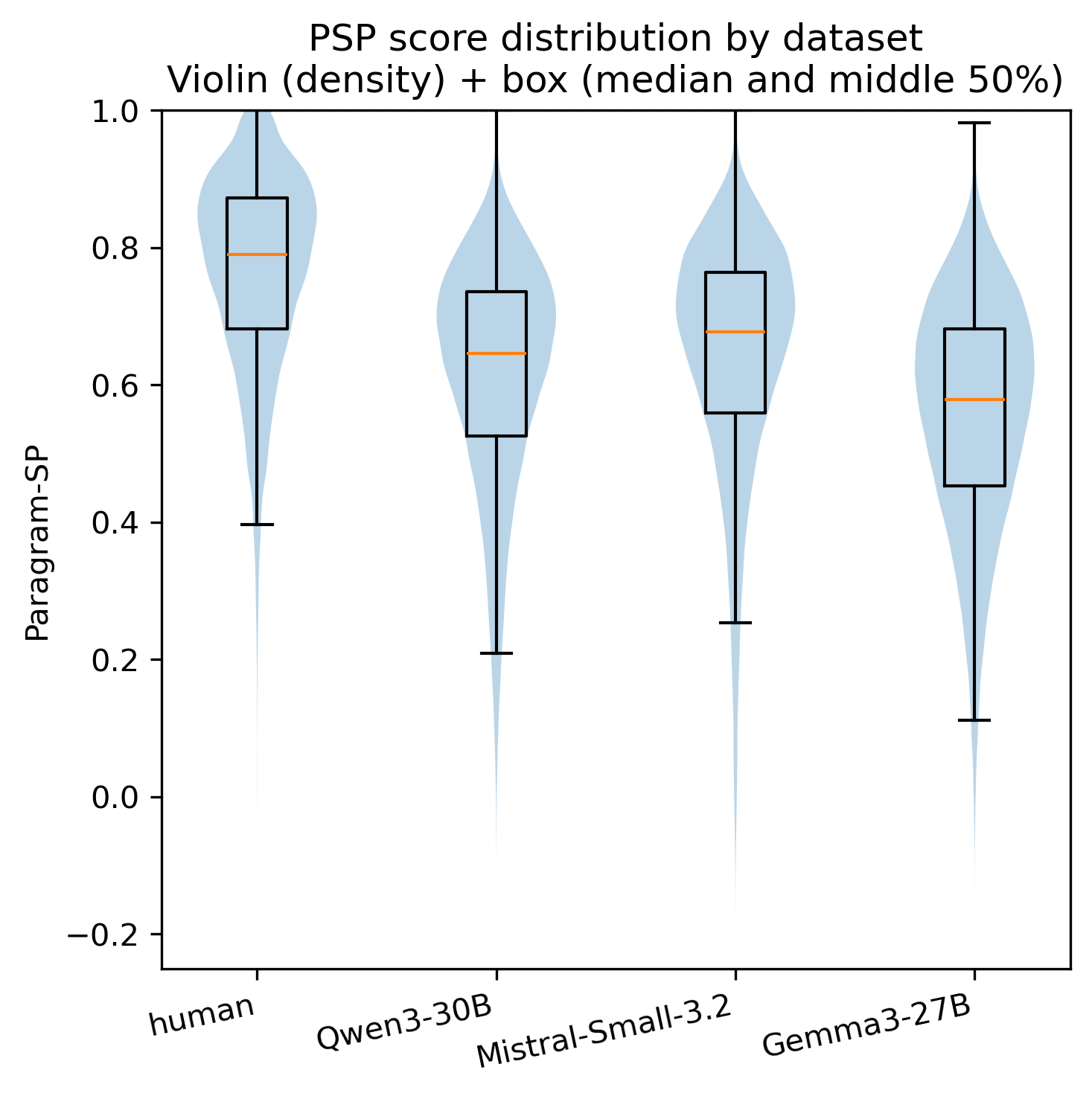}
\includegraphics[width=0.49\linewidth]{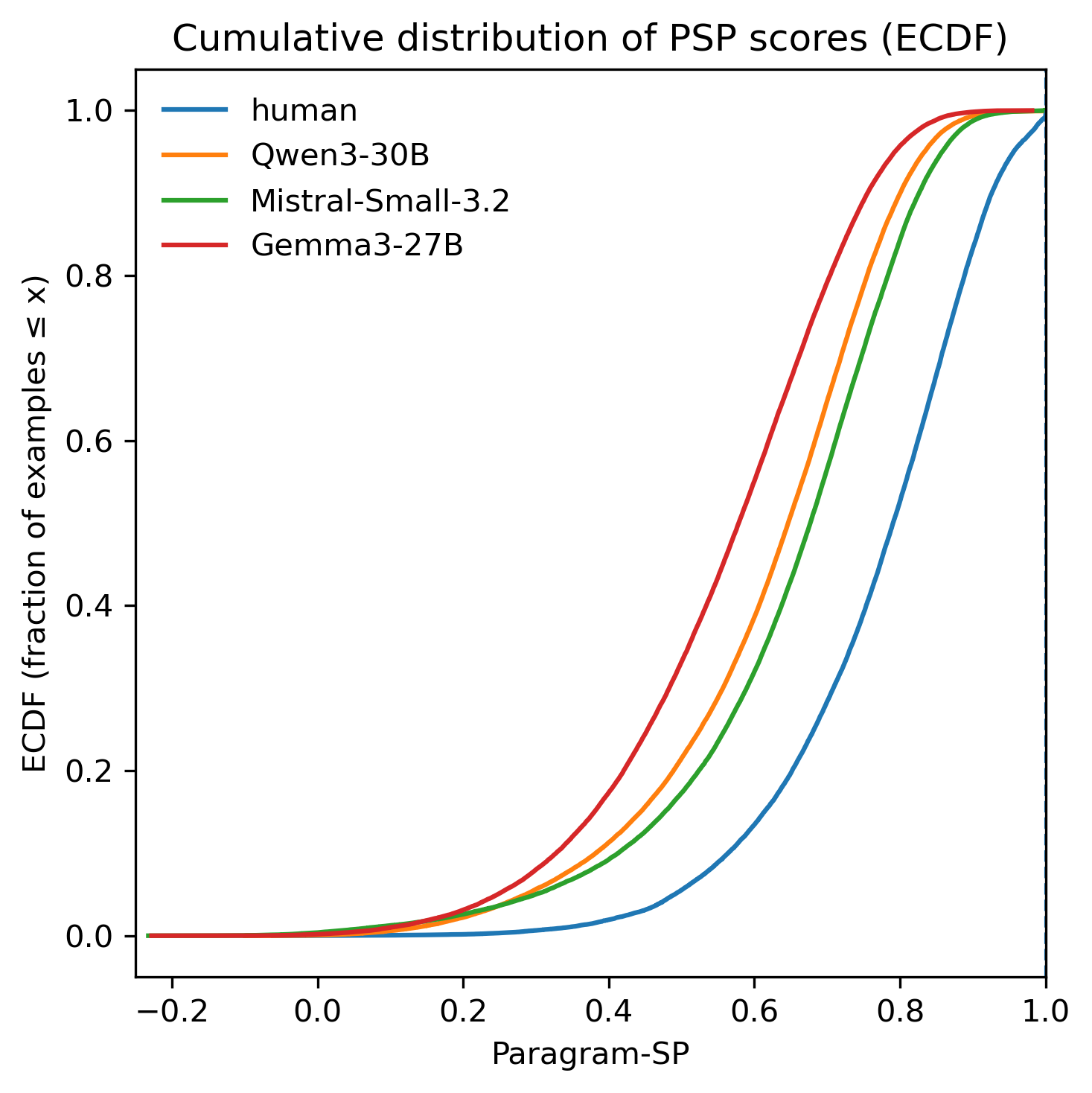}
\caption{The distributions of the pair-wise semantic similarity score between one generated additional reference and the human translation. The top row is from the French-English data, and bottom row is from the Russian-English data. ``Human'' pair-wise similarity score distribution is from the multi-reference full medium+high \semsim dataset. Overall, the synthetic pairwise similarity score distribution shifts lower compared to the human translation distribution.}
\label{fig-syn-simp-violin-ecdf}
\end{figure}

In Figure~\ref{fig-syn-simp-violin-ecdf}, we compare the distribution of the pairwise \spavgfull scores between the ground-truth human translation and each LLM's zero-shot paraphrased translation.

Compared to the \spavgfull scores between two human translations, the score between a human and a synthetic translation is on average lower across all models.

In the violin plot, the middle 50\% of the distribution shifts down for all three models, as does the median. This means the \semsim of synthetically augmented data is lower overall, since \semsim is the average pairwise score among all references of the same source. Interestingly, Gemma3 deviates the most from the human score distribution, yet its paraphrases gave the best MT performance of the three models. Note that for all three models, the majority of the data falls in the medium \semsim range, whereas nearly half of the human data falls in the high \semsim range.

\newpage
\section{Prompt for Classifying Translators' comments} \label{appendix-classify-translator-comment}
We used Claude Opus 4.6 to label annotator comments with fine-grained translation quality categories. For each comment, the model was given the original French source text, Translation A, Translation B, the annotator's free-text comment, and the full list of quality labels. We wrote this detailed instruction with the assistance of Claude Opus 4.6 in the chat interface. We tested multiple instructions by checking 50 outputs to ensure accuracy. Below is the prompt we use:

\begin{quote}
\small
You are a strict annotation analyst for literary translation evaluation (French to English). You will be given:
\begin{itemize}
    \item The original French source text
    \item Translation A and Translation B (the actual translations)
    \item A human annotator's comment comparing them
    \item A list of quality labels
\end{itemize}

For EACH quality label, decide:
\begin{itemize}
    \item ``A'' if the comment EXPLICITLY praises or criticizes Translation A for this quality
    \item ``B'' if the comment EXPLICITLY praises or criticizes Translation B for this quality
    \item ``tie'' if the comment EXPLICITLY says BOTH translations share this quality
    \item ``'' (empty) if the comment does NOT clearly address this quality
\end{itemize}

STRICT RULES:
\begin{itemize}
    \item Only tag a quality if the comment clearly and directly discusses it. Do NOT infer qualities that are merely implied.
    \item Most qualities should be ``'' (empty) for any given comment. A typical comment only addresses 1--3 qualities.
    \item Annotators often quote specific words or phrases from the translations without saying ``A'' or ``B''. You MUST match quoted words back to Translation A or Translation B to determine which system they refer to.
    \item If the comment says one translation ``flows better'' or ``is more natural'', that is ``fluent / natural'' for that system AND ``awkward / unnatural'' for the other ONLY if the comment indicates the other is worse.
    \item Do NOT automatically assign the opposite negative quality just because one system got a positive. Only assign it if the comment actually criticizes the other.
\end{itemize}

Quality definitions:
\begin{itemize}
    \item ``fluent / natural'': reads smoothly, sounds like natural English
    \item ``accurate / faithful'': correctly conveys the meaning of the French source
    \item ``idiomatic / good word choice'': uses appropriate English expressions and well-chosen words
    \item ``good style / tone'': preserves the literary register, mood, or voice of the original
    \item ``awkward / unnatural'': reads clumsily, sounds unnatural or stilted
    \item ``overly literal'': follows French syntax or phrasing too closely, word-for-word translation
    \item ``mistranslation / meaning error'': gets the meaning wrong, misinterprets the source
    \item ``omission / addition'': leaves out content from the source or adds content not in the source
    \item ``grammar / mechanical error'': has grammatical mistakes, typos, or mechanical issues
\end{itemize}

EXAMPLES:

Example 1:

Original French: [French source text]

Translation A: [some translation with more natural English]

Translation B: [some translation with less natural English]

Comment: ``I've gone with A purely because the English is more natural and flows more smoothly than B. However, there are serious issues in both: neither gets d'Artagnan's name right, neither explains what a conge is (it would have been better to write `permissions to leave' or something like that and I'd have preferred `pistols' to `pistoles'.''

Correct labels:

\{``fluent / natural'': ``A'', ``accurate / faithful'': ``'', ``idiomatic / good word choice'': ``'', ``good style / tone'': ``'', ``awkward / unnatural'': ``B'', ``overly literal'': ``'', ``mistranslation / meaning error'': ``tie'', ``omission / addition'': ``'', ``grammar / mechanical error'': ``''\}

Example 2:

Original French: [French source text]

Translation A: [some translation preserving Milady's voice]

Translation B: [some translation]

Comment: ``A sounds more like Milady would speak.''

Correct labels:

\{``fluent / natural'': ``'', ``accurate / faithful'': ``'', ``idiomatic / good word choice'': ``'', ``good style / tone'': ``A'', ``awkward / unnatural'': ``'', ``overly literal'': ``'', ``mistranslation / meaning error'': ``'', ``omission / addition'': ``'', ``grammar / mechanical error'': ``''\}

Respond with ONLY a JSON object. No explanation, no markdown, no backticks.
\end{quote}

For each instance, the user prompt was constructed using the following template:

\begin{quote}
\small
Original French: \{original\}

Translation A: \{a\_pred\}

Translation B: \{b\_pred\}

Annotator comment:

``\{comment\}''

Quality labels:

- fluent / natural

- accurate / faithful

- idiomatic / good word choice

- good style / tone

- awkward / unnatural

- overly literal

- mistranslation / meaning error

- omission / addition

- grammar / mechanical error

Return a JSON object mapping each quality label to ``A'', ``B'', ``tie'', or ``''. Remember: most should be empty. Only tag what the comment explicitly discusses.
\end{quote}

\section{Descriptions of Each \semsim Level (Table \ref{semantic_sim_table})} \label{Appendix-semsim-description} 

\begin{table*}[h]\centering
\small
\begin{tabular}{l l l}
\toprule
   \textbf{\semsim}               & \textbf{Description}  & \textbf{Dataset}\\
 \midrule
$1.0$             &   Completely identical.      & Medium+High     \\
                  &                                                   & Unfiltered  \\
\midrule
$[0.9-1.0)$  &  Almost identical, except for minor punctuation differences,  & Medium+High\\ 
             &  e.g., single vs double quotation marks, or word replacement & Unfiltered\\
             & (by a synonym or a phrase). & \\
              \midrule
$[0.85-0.9)$ & Differences happen more often, but they are still variations   & Medium+High\\
             & that are not meaningful. Compared to the 0.9-1.0 similarity,  &  Unfiltered\\
             & the difference can be clausal instead of just one or two words.   &\\
             & However, the overall amount of information is about the same. &\\
             \midrule
$[0.45-0.85)$  &  Very meaningful variations. The paraphrases are more likely  & Medium\\
               & to be visually dissimilar. Sometimes, the amount of information &Medium+High\\
               & can differ much more: some translators are more verbose,    & Medium+Low\\
               &  others are more succinct. & Unfiltered\\
                \midrule
$[-1-0.45)$  &    Noisy paraphrases that have an overall information mismatch.  & Medium+Low\\
            &  Paraphrases can be misaligned or erroneous, or translators & Unfiltered\\
            &   diverge too much in their interpretations. &\\
\bottomrule
\end{tabular}
\caption{Different levels of semantic similarity score (\semsim) and their characteristics. See Section \ref{subsec-semsim} for how we compute \semsim given a source text and its reference translations. The dataset column indicates the datasets that include paraphrases from the corresponding \semsim range.}
\label{semantic_sim_table}
\end{table*}

\newpage 
\section{Human Evaluation Interface Details} \label{appendix-human-eval}

Across all tasks, no source text or translation is repeated within a survey, across survey variations for the same task, or in any other task. 

Translators were informed of the duration, the pay, and the nature of the task before accepting the job. 

\begin{figure*}[h]
\centering
\includegraphics[width=0.9\linewidth]{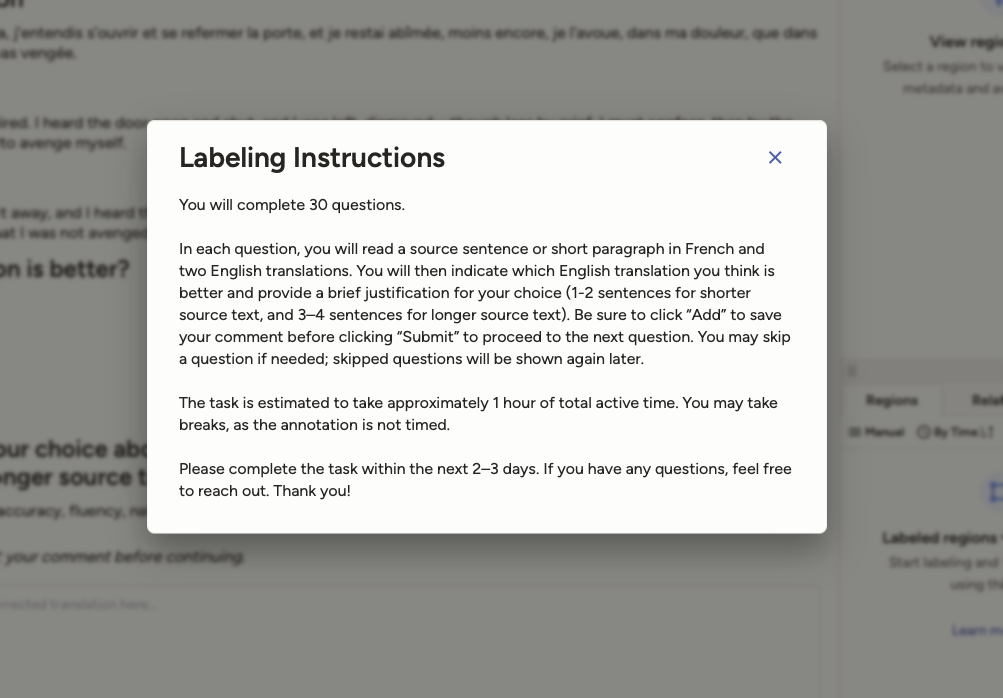}
\caption{The instructions shown to translators before starting the annotation process. The human evaluation task was hosted on Label Studio.}
\label{fig-anno-instruct}
\end{figure*}

\begin{figure*}[h]
\centering
\includegraphics[width=0.9\linewidth]{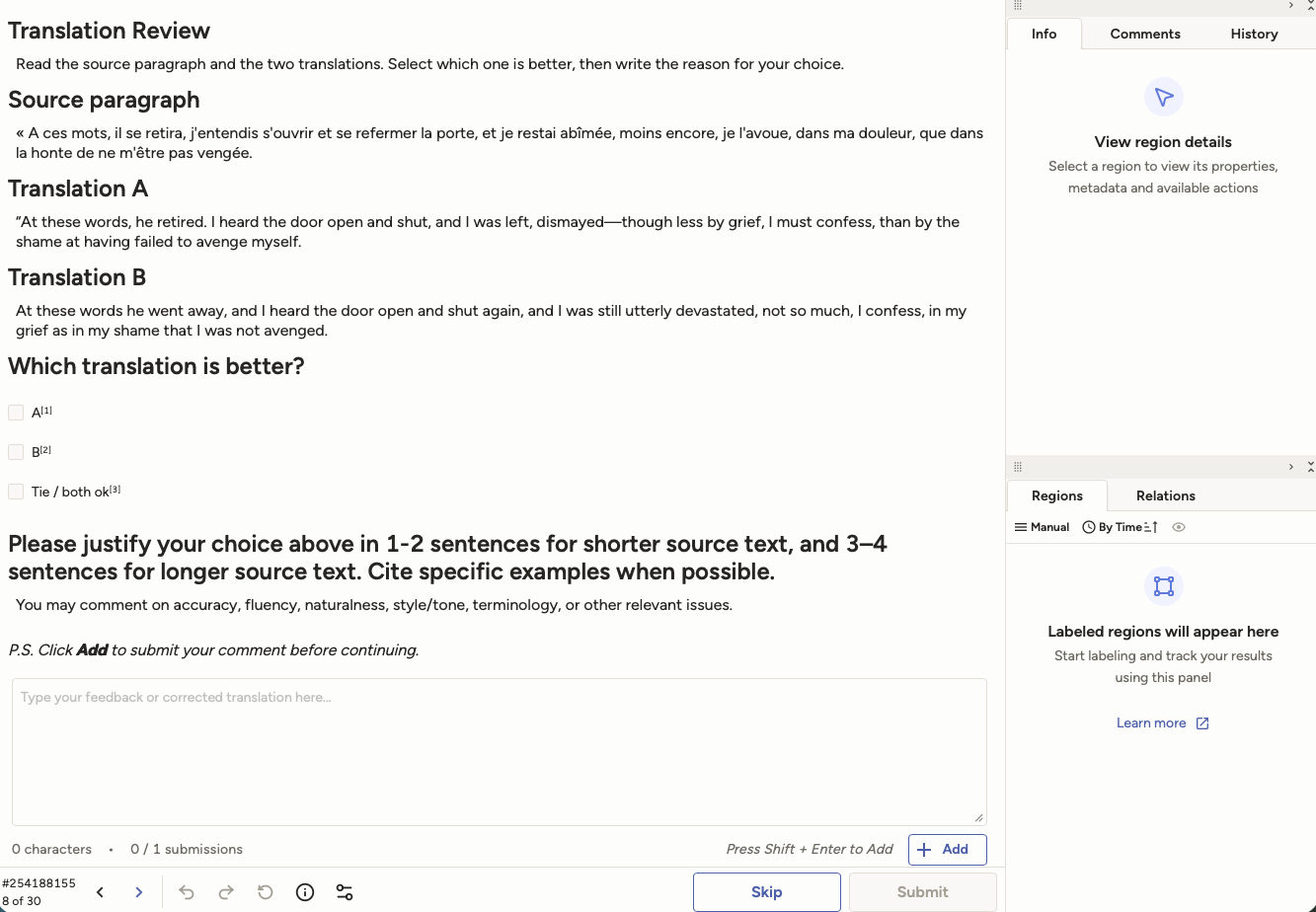}
\caption{A screenshot of the Label Studio interface used for human evaluation, showing an example question presented to translators.}
\label{fig-anno-ui}
\end{figure*}

In Figure~\ref{fig-anno-instruct} and \ref{fig-anno-ui}, we show the instructions and the UI for annotation on Label Studio, as well as an example question. 

\section{Example Translation Outputs from Different MT Systems (Table \ref{tab:example-outputs})}
\label{appendix-example-outputs}

\begin{table*}[t]
\centering
\small
\centering
\renewcommand{\arraystretch}{1.3}
\begin{tabularx}{\textwidth}{p{2.2cm}X}
\toprule
\textbf{System} & \textbf{Output} \\
\midrule
Source (FR) & \textit{— Ah! voilà qui est nouveau, reprit le roi. N'allez-vous pas dire que vos trois mousquetaires damnés, Athos, Porthos et Aramis et votre cadet de Béarn, ne se sont pas jetés comme des furieux sur le pauvre Bernajoux, et ne l'ont pas maltraité de telle façon qu'il est probable qu'il est en train de trépasser à cette heure! N'allez-vous pas dire qu'ensuite ils n'ont pas fait le siège de l'hôtel du duc de La Trémouille, et qu'ils n'ont point voulu le brûler! ce qui n'aurait peut-être pas été un très grand malheur en temps de guerre, vu que c'est un nid de huguenots, mais ce qui, en temps de paix, est un fâcheux exemple. Dites, n'allez-vous pas nier tout cela?} \\
\midrule
Human 1 & ``Ah! That's a new one,'' replied the king. ``Are you going to tell me that your three damned musketeers, Athos, Porthos, and Aramis, and your cadet from B\'earn, haven't sprung like furies on poor Bernajoux and injured him so badly that by now he's probably dead? Are you going to tell me that they didn't lay siege to the h\^otel of the Duc de La Tr\'emouille, and that they didn't try to burn it down? Of course, that might not be such a bad thing in times of war, since it's a nest of Huguenots---but in peacetime, it sets a bad example! Speak! Isn't this what you're going to tell me?'' \\
\midrule
Human 2 & ``Ah! here's something new!'' the king picked up. ``Are you going to tell me that your three damned musketeers, Athos, Porthos, and Aramis, and your lad from B\'earn, didn't throw themselves like madmen on poor Bernajoux and mistreat him so badly that he's probably on his way out of this world as I speak? Are you going to say that after that they didn't lay siege to the h\^otel of the duc de La Tr\'emouille, and that they didn't intend to burn it? Which would perhaps have been no great misfortune in time of war, seeing that it's a nest of Huguenots, but which, in time of peace, is a regrettable example. Tell me, are you going to deny all that?'' \\
\midrule
Unfiltered (all) & ``Ah! this is new,'' resumed the king. ``Do you not say that your three damned musketeers, Athens, Porthos, and Aramis, and your Cadet de B\'earn, did not rush like madmen upon poor Bernajoux, and did not so badly treat him that he is probably dying at this hour? Do you not say that they did not siege the Duke de La Tr\'emouille's house, and did not want to burn him! which would not have been a great misfortune in war, since it is a nest of Huguenots, but which in peace is a dreadful example. Say, do you not deny all this?'' \\
\midrule
Medium+high (all) & ``Ah! this is something new,” continued the king. “Don’t you say that your three damned musket-boys, Athenos, Porthos, and Aramis, and your Cadet de B\'earn, did not throw themselves like mad on poor Bernajoux, and not so ill-treated him that he is probably in the process of dying at this hour! Don’t you say that they did not seize the Duke de La Tr\'emouille's house, and that they did not want to burn him! This would perhaps not have been a great misfortune in war, since it is the nest of Huguenots, but which in peace is a nasty example. Say, will you not deny all this?'' \\
\midrule
Synthetic (Gemma3) & ``Ah! that is new,'' resumed the king. ``Don't you say that your three damned mousquetaires, Athens, Porthos, Aramis, and your cadet of B\'earn, did not throw themselves like furious beasts upon poor Bernouch, and did not treat him so badly that he is probably on the point of passing through! Don't you say that they did not take siege on the Duke de La Tr\'emouille's house, and that they did not want to burn him! which would not have been very bad in wartime, since it was a nest of Hunchbacks, but which, in peace, is a nasty example. Say, will you not deny all this?'' \\
\midrule
Synthetic (Mistral) & ``Ah, this is a new thing,'' resumed the king. ``Don't you say that your three damned musketeers, Athens, Porthos, and Aramis, and your scholar of B\'earn, did not throw themselves like furious beasts upon poor Bernajoux, and did not treat him so badly that he is probably in the process of passing this hour! Don't you say that they did not seize the siege of the Duke de La Tr\'emouille's house, and did not want to burn him! which would perhaps not have been a great misfortune in war, since it is a nest of Huguenots, but which, in peace, is a dreadful example. Say, will you not deny all this?'' \\
\midrule
Synthetic (Qwen3) & `Ah, that's nothing new,' the king went on. `Don't you say that your three damned spies, Athenaeus, Porthos and Arimas, and your cadet of Benares, did not throw themselves like a madman upon poor Bernard, and did not treat him so badly that he is probably going to be passing through this hour! Don't you say that they did not take the Duke of La Tr\'emo\"illes' house, and did not want to burn it! What would have been a great misfortune in war, since it was a nest of Hunchbacks, but what is a shameful example in peace. Say, won't you hide all this?' \\
\bottomrule
\end{tabularx}

\caption{Example system outputs for a source paragraph from \textit{Les Trois Mousquetaires} (\textit{The Three Musketeers}) by Alexandre Dumas.}
\label{tab:example-outputs}
\end{table*}

\end{document}